%% file: main.tex
\documentclass[10pt,twocolumn,letterpaper]{article}

\usepackage{cvpr}              

\input{preamble}

\definecolor{cvprblue}{rgb}{0.21,0.49,0.74}
\usepackage[pagebackref,breaklinks,colorlinks,urlcolor=cvprblue,linkcolor=cvprblue,citecolor=cvprblue]{hyperref}
\usepackage{inconsolata}
\usepackage{multirow}
\usepackage{tabularx}
\usepackage{colortbl}
\usepackage{pifont}
\usepackage{multicol}
\usepackage{xcolor}
\def\paperID{742} 
\def\confName{CVPR}
\def\confYear{2024}

\usepackage{colortbl}
\usepackage{xcolor}
\definecolor{tabhighlight}{HTML}{e5e5e5}

\definecolor{mydarkgreen}{rgb}{0.0, 0.5, 0.0}  

\usepackage{color}
\usepackage{marvosym}
\usepackage{enumitem}
\usepackage{utfsym}
\usepackage{xspace}
\usepackage{pifont}
\usepackage{arydshln}
\usepackage[T1]{fontenc}
\definecolor{citecolor}{HTML}{0071BC}
\definecolor{linkcolor}{HTML}{ED1C24}
\definecolor{LGray}{gray}{0.97}

\usepackage{wrapfig}
\newcommand{\tableCellHeight}{1}
\newcommand{\tabstyle}[1]{
  \setlength{\tabcolsep}{#1}
  \renewcommand{\arraystretch}{\tableCellHeight}
  \centering
  \small
}

\usepackage{sidecap}
\usepackage{booktabs}
\renewcommand{\arraystretch}{1}
\usepackage{afterpage}

\title{GLaMM: Pixel Grounding Large Multimodal Model}
\author{%
  Hanoona Rasheed$^{1}$\textsuperscript{\textnormal{*}},~~Muhammad Maaz$^{1}$\textsuperscript{\textnormal{*}},~~Sahal Shaji$^{1}$, Abdelrahman Shaker$^{1}$,~~Salman Khan$^{1,2}$ \\~~Hisham Cholakkal$^{1}$,  Rao M. Anwer$^{1,3}$,~~Eric Xing$^{1,4}$,~~Ming-Hsuan Yang$^{5,7}$,~~Fahad S. Khan$^{1,6}$\\[0.25cm]
 {\fontsize{10.5pt}{12pt}\selectfont $^{1}$Mohamed bin Zayed University of AI, $^{2}$Australian National University, $^{3}$Aalto University}\\
 {\fontsize{10.5pt}{12pt}\selectfont $^{4}$Carnegie Mellon University, $^{5}$University of California - Merced, $^{6}$Linköping University, $^{7}$Google Research}\\
  \hypersetup{urlcolor=black}
  {\normalsize \href{mailto:hanoona.bangalath@mbzuai.ac.ae}{hanoona.bangalath@mbzuai.ac.ae}, \href{mailto:muhammad.maaz@mbzuai.ac.ae}{muhammad.maaz@mbzuai.ac.ae} }\\
{\hypersetup{urlcolor=golden}
  \fontsize{10pt}{12pt}\selectfont \href{https://github.com/mbzuai-oryx/groundingLMM}{https://github.com/mbzuai-oryx/groundingLMM}, \href{https://grounding-anything.com/}{https://grounding-anything.com}}
}

\begin{document}
\maketitle

\input{sec/0_abstract}    
\input{sec/1_intro}

\input{sec/2_related}

\input{sec/4_method}
\input{sec/3_dataset}

\input{sec/5_experiments}
\input{sec/conclusion}
{
    \small
    \bibliographystyle{ieeenat_fullname}
    \bibliography{main}
}
\clearpage
\input{sec/X_suppl}

\end{document}

%% file: preamble.tex
\usepackage[dvipsnames]{xcolor}

\usepackage{xcolor}
\definecolor{golden}{RGB}{171, 107, 35}

%% file: sec/0_abstract.tex
\begin{abstract}
Large Multimodal Models (LMMs) extend Large Language Models to the vision domain. Initial LMMs used holistic images and text prompts to generate ungrounded textual responses. 
Recently, region-level LMMs have been used to generate visually grounded responses. However, they are limited to only referring to a single object category at a time, require users to specify the regions,
or cannot offer dense pixel-wise object grounding. In this work, we present Grounding LMM (GLaMM), the first model that can generate natural language responses seamlessly intertwined with corresponding object segmentation masks. GLaMM not only grounds objects appearing in the conversations but is flexible enough to accept both textual and optional visual prompts (region of interest) as input. This empowers users to interact with the model at various levels of granularity, both in textual and visual domains. Due to the lack of standard benchmarks for the novel setting of visually Grounded Conversation Generation (GCG), we introduce a comprehensive evaluation protocol with our curated grounded conversations. Our proposed GCG task requires densely grounded concepts in natural scenes at a large-scale. To this end, we propose a densely annotated Grounding-anything Dataset (GranD) using our proposed automated annotation pipeline that encompasses 7.5M unique concepts grounded in a total of 810M regions available with segmentation masks. Besides GCG, GLaMM also performs effectively on several downstream tasks, e.g., referring expression segmentation, image and region-level captioning and vision-language conversations.
\footnotetext[1]{Equal contribution.}
\end{abstract}

%% file: sec/1_intro.tex
\begin{figure}[h!tp]
  \centering
    \includegraphics[width=\linewidth]{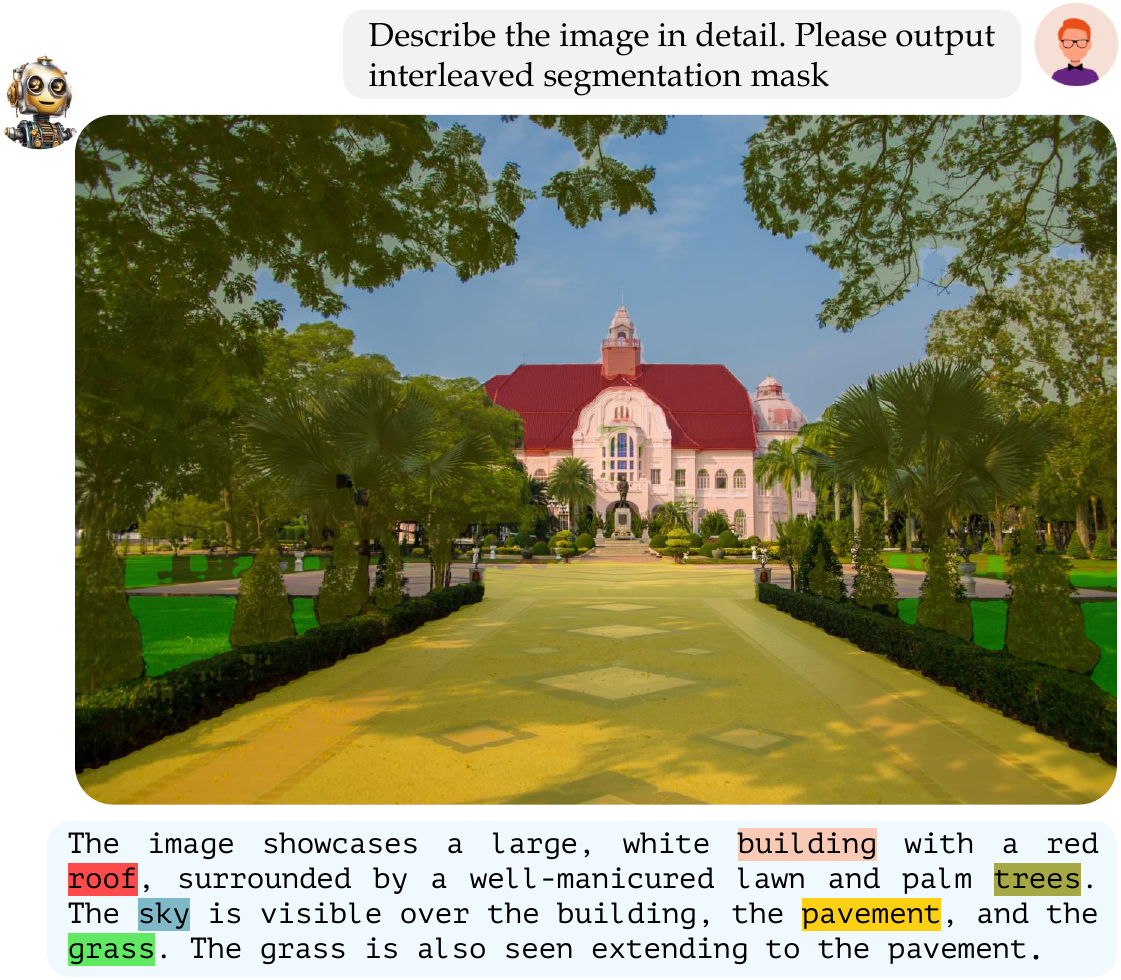}\vspace{-0.5em}
    \caption{\textbf{Grounded Conversation Generation with GLaMM}. Our multimodal conversational model can provide natural language responses grounded at the pixel-level in the input image. Different levels of granularity are depicted in the output groundings, e.g., things (\emph{building}, \emph{tree}), stuff (\emph{grass}, \emph{sky}, \emph{pavement}), and object parts (\emph{roof} as a subpart of the building) alongside the object attributes (\emph{white house}, \emph{red roof}, \emph{well-manicured lawn}) and object relationships (\emph{grass extending to the pavement}, \emph{sky over the building}). Existing LMMs, open-source (e.g., LLaVa, miniGPT4, Shikra, Kosmos-2) and closed-source (e.g., GPT4-V, Bard), do not offer pixel-level grounded conversational capability. }
    \label{fig:fig1}
\vspace{-1.5em}
\end{figure}

\vspace{-1.9em}
\section{Introduction}
\label{sec:intro}
Fueled by the generative AI wave, Large Multimodal Models (LMMs) have emerged as a pivotal advancement, bridging the gap between vision and language tasks \cite{awais2023foundational}. Initial efforts like \cite{liu2023llava, zhu2023minigpt, li2023otter, dai2023instructblip, gao2023llamaadapterv2, ye2023mplugowl} demonstrate effective textual responses based on input images. Although these models are sophisticated, they cannot still ground their responses in the visual context. Such grounding is crucial for advanced applications like detailed visual understanding, interactive embodied agents, and localized content manipulation. Recent efforts have started to address this limitation by enabling models to process user-defined regions specified via bounding boxes~\cite{zhang2023gpt4roi, liu2023interngpt, pi2023detgpt, peng2023kosmos, chen2023shikra}.

A few recent works have explored grounded text response generation~\cite{lai2023lisa, zhao2023bubogpt, peng2023kosmos, chen2023shikra}
but do not provide detailed \emph{pixel-level} groundings. Parallel to these, efforts have been made in the referring segmentation literature to ground textual descriptions in natural images~\cite{lai2023lisa}. However, they are limited to grounding a single object and cannot engage in natural, coherent \emph{conversations}, thereby restricting their practical applicability in interactive tasks that demand a deep understanding of both visual and textual content. To address these limitations of existing works, we introduce \textit{Grounding LMM} (GLaMM), that simultaneously provides in-depth region understanding, pixel-level groundings, and conversational abilities through an end-to-end training approach (see Fig.~\ref{fig:fig1} and Tab.~\ref{tab:methods_comparison}).

To address the lack of benchmarks for visually grounded conversations, we introduce the novel task of \textit{Grounded Conversation Generation} (GCG). The GCG task aims to produce natural language responses interleaved with object segmentation masks. This challenging task unifies several existing tasks in computer vision that are typically treated in isolation, i.e., referring expression segmentation, image and region-level captioning, phrase grounding, and vision-language conversations. Thereby, our unified model and proposed pretraining dataset can effectively transfer to several downstream tasks (referring expression segmentation, region-level captioning, image captioning, and conversational-style QA).  We present GLaMM as the first model specifically designed for this challenging task. Unlike prior works, GLaMM can work with both textual and visual prompts and can generate visually grounded outputs, thus offering a versatile user experience.

Detailed region-level understanding requires the laborious process of collecting large-scale annotations for image regions. We propose an automated pipeline to annotate the large-scale \textit{Grounding-anything Dataset} (GranD) to alleviate the manual labeling effort. Leveraging the automated pipeline with dedicated verification steps, GranD comprises 7.5M unique concepts anchored in 810M regions, each with a segmentation mask. Using state-of-the-art vision and language models, the dataset annotates SAM~\cite{kirillov2023segment} images through a multi-level hierarchical scheme that enhances annotation quality. With 11M images, 84M referring expressions, and 33M grounded captions, GranD sets a new benchmark in comprehensiveness. In addition to the automatically generated dataset for the GCG, we provide the first high-quality dataset for grounded conversations obtained by revamping the existing manually annotated datasets~\cite{plummer2015flickr30k, kazemzadeh2014referitgame, yang2022psg} for GCG using GPT-4~\cite{openai2023gpt4} in-context learning. We refer to the high-quality dataset as GranD$_f$, denoting its suitability for fine-tuning. 

Our work has three main contributions:
\begin{itemize}
    \item We present GLaMM, the first model capable of generating natural language responses seamlessly integrated with object segmentation masks. Unlike existing models, GLaMM accommodates textual and visual prompts, facilitating enhanced multimodal user interaction.
    \item Recognizing the lack of standardized benchmarks for visually grounded conversations, we propose the new Grounded Conversation Generation (GCG) task. We also introduce a comprehensive evaluation protocol to measure the efficacy of models for GCG that unifies multiple isolated tasks, filling a significant gap in the literature.
    \item To facilitate model training and evaluation, we create Grounding-anything Dataset (GranD), a large-scale densely annotated dataset. Developed using an automatic annotation pipeline and verification criteria, it encompasses 7.5M unique concepts grounded in 810M regions. Additionally, we propose GranD$_f$, a high-quality dataset explicitly designed for the GCG task finetuning, by re-purposing existing open-source datasets.
    
\end{itemize}

\begin{table*}[!t]
    \centering
    \resizebox{0.98\textwidth}{!}{%
    \begin{tabular}{lccccccc}
        \toprule
       \multirow{2}{*}{Method} & \multirow{2}{*}{Image} & \multicolumn{2}{c}{Input / Output} & \multirow{2}{*}{Region} & \multirow{2}{*}{Pixel-Wise} & \multirow{2}{*}{Multi-turn} & \multirow{2}{*}{End-End} \\
        \cmidrule{3-4} 
        & & Region & Multi-Region & Enc. / Dec. & Grounding & Conversation & Model \\
        \midrule

    \rowcolor{LGray} MM-REACT (arXiv-23)~\cite{yang2023mmreact} & \textcolor{ForestGreen}{\usym{2713}} & \textcolor{red}{\usym{2717}}~/~\textcolor{red}{\usym{2717}} & \textcolor{red}{\usym{2717}}~/~\textcolor{red}{\usym{2717}} & \textcolor{red}{\usym{2717}}~/~\textcolor{red}{\usym{2717}} & \textcolor{red}{\usym{2717}} & \textcolor{ForestGreen}{\usym{2713}} & \textcolor{red}{\usym{2717}} \\
    
    LLaVA (NeurIPS-23)~\cite{liu2023llava} & \textcolor{ForestGreen}{\usym{2713}} & \textcolor{red}{\usym{2717}}~/~\textcolor{red}{\usym{2717}} & \textcolor{red}{\usym{2717}}~/~\textcolor{red}{\usym{2717}} & \textcolor{red}{\usym{2717}}~/~\textcolor{red}{\usym{2717}} & \textcolor{red}{\usym{2717}} & \textcolor{ForestGreen}{\usym{2713}} & \textcolor{ForestGreen}{\usym{2713}} \\
    
    \rowcolor{LGray} miniGPT4 (arXiv-23)~\cite{zhu2023minigpt} & \textcolor{ForestGreen}{\usym{2713}} & \textcolor{red}{\usym{2717}}~/~\textcolor{red}{\usym{2717}} & \textcolor{red}{\usym{2717}}~/~\textcolor{red}{\usym{2717}} & \textcolor{red}{\usym{2717}}~/~\textcolor{red}{\usym{2717}} & \textcolor{red}{\usym{2717}} & \textcolor{ForestGreen}{\usym{2713}} & \textcolor{ForestGreen}{\usym{2713}} \\

    mPLUG-OWL (arXiv-23)~\cite{ye2023mplugowl} & \textcolor{ForestGreen}{\usym{2713}} & \textcolor{red}{\usym{2717}}~/~\textcolor{red}{\usym{2717}} & \textcolor{red}{\usym{2717}}~/~\textcolor{red}{\usym{2717}} & \textcolor{red}{\usym{2717}}~/~\textcolor{red}{\usym{2717}} & \textcolor{red}{\usym{2717}} & \textcolor{ForestGreen}{\usym{2713}} & \textcolor{ForestGreen}{\usym{2713}} \\
    
    \rowcolor{LGray} LLaMA-Adapter v2 (arXiv-23)~\cite{gao2023llamaadapterv2} & \textcolor{ForestGreen}{\usym{2713}} & \textcolor{red}{\usym{2717}}~/~\textcolor{red}{\usym{2717}} & \textcolor{red}{\usym{2717}}~/~\textcolor{red}{\usym{2717}} & \textcolor{red}{\usym{2717}}~/~\textcolor{red}{\usym{2717}} & \textcolor{red}{\usym{2717}} & \textcolor{ForestGreen}{\usym{2713}} & \textcolor{ForestGreen}{\usym{2713}} \\
    
    Otter (arXiv-23)~\cite{li2023otter} & \textcolor{ForestGreen}{\usym{2713}} & \textcolor{red}{\usym{2717}}~/~\textcolor{red}{\usym{2717}} & \textcolor{red}{\usym{2717}}~/~\textcolor{red}{\usym{2717}} & \textcolor{red}{\usym{2717}}~/~\textcolor{red}{\usym{2717}} & \textcolor{red}{\usym{2717}} & ~\textcolor{red}{\usym{2717}} & \textcolor{ForestGreen}{\usym{2713}} \\
    
    \rowcolor{LGray} Instruct-BLIP (arXiv-23)~\cite{dai2023instructblip} & \textcolor{ForestGreen}{\usym{2713}} & \textcolor{red}{\usym{2717}}~/~\textcolor{red}{\usym{2717}} & \textcolor{red}{\usym{2717}}~/~\textcolor{red}{\usym{2717}} & \textcolor{red}{\usym{2717}}~/~\textcolor{red}{\usym{2717}} & \textcolor{red}{\usym{2717}} & \textcolor{ForestGreen}{\usym{2713}} & \textcolor{ForestGreen}{\usym{2713}} \\

    \arrayrulecolor{black}
    \cdashline{1-8}[1.5pt/4pt]
    
    InternGPT (arXiv-23)~\cite{liu2023interngpt} & \textcolor{ForestGreen}{\usym{2713}} & \textcolor{ForestGreen}{\usym{2713}}~/~\textcolor{red}{\usym{2717}} & \textcolor{red}{\usym{2717}}~/~\textcolor{red}{\usym{2717}} & \textcolor{red}{\usym{2717}}~/~\textcolor{red}{\usym{2717}} & \textcolor{red}{\usym{2717}} & \textcolor{ForestGreen}{\usym{2713}} & \textcolor{red}{\usym{2717}} \\

    \rowcolor{LGray} Bubo-GPT (arXiv-23)~\cite{zhao2023bubogpt} & \textcolor{ForestGreen}{\usym{2713}} & \textcolor{red}{\usym{2717}}~/\textcolor{ForestGreen}{\usym{2713}} & \textcolor{red}{\usym{2717}}~/\textcolor{ForestGreen}{\usym{2713}} & \textcolor{red}{\usym{2717}}~/~\textcolor{red}{\usym{2717}} & \textcolor{red}{\usym{2717}} & \textcolor{ForestGreen}{\usym{2713}} & \textcolor{red}{\usym{2717}} \\

    \arrayrulecolor{black}
    \cdashline{1-8}[1.5pt/4pt]

    Vision-LLM (arXiv-23)~\cite{wang2023visionllm} & \textcolor{ForestGreen}{\usym{2713}} & \textcolor{red}{\usym{2717}}~/~\textcolor{ForestGreen}{\usym{2713}} & \textcolor{red}{\usym{2717}}~/~\textcolor{ForestGreen}{\usym{2713}} & \textcolor{red}{\usym{2717}}~/~\textcolor{red}{\usym{2717}} & \textcolor{red}{\usym{2717}} & \textcolor{red}{\usym{2717}} & \textcolor{ForestGreen}{\usym{2713}} \\
    
    \rowcolor{LGray} Det-GPT (arXiv-23)~\cite{pi2023detgpt} & \textcolor{ForestGreen}{\usym{2713}} & \textcolor{ForestGreen}{\usym{2713}}~/~\textcolor{ForestGreen}{\usym{2713}} & \textcolor{ForestGreen}{\usym{2713}}~/~\textcolor{ForestGreen}{\usym{2713}} & \textcolor{red}{\usym{2717}}~/~\textcolor{red}{\usym{2717}} & \textcolor{red}{\usym{2717}} & \textcolor{ForestGreen}{\usym{2713}} & \textcolor{ForestGreen}{\usym{2713}} \\
    
    Shikra (arXiv-23)~\cite{chen2023shikra} & \textcolor{ForestGreen}{\usym{2713}} & \textcolor{ForestGreen}{\usym{2713}}~/~\textcolor{ForestGreen}{\usym{2713}} & \textcolor{red}{\usym{2717}}~/~\textcolor{red}{\usym{2717}} & \textcolor{red}{\usym{2717}}~/~\textcolor{red}{\usym{2717}} & \textcolor{red}{\usym{2717}} & \textcolor{red}{\usym{2717}} & \textcolor{ForestGreen}{\usym{2713}} \\
    
    \rowcolor{LGray} Kosmos-2 (arXiv-23)~\cite{peng2023kosmos} & \textcolor{ForestGreen}{\usym{2713}} & \textcolor{ForestGreen}{\usym{2713}}~/~\textcolor{ForestGreen}{\usym{2713}} & \textcolor{ForestGreen}{\usym{2713}}~/~\textcolor{ForestGreen}{\usym{2713}} & \textcolor{red}{\usym{2717}}~/~\textcolor{red}{\usym{2717}} & \textcolor{red}{\usym{2717}} & \textcolor{red}{\usym{2717}} & \textcolor{ForestGreen}{\usym{2713}} \\
    
    \arrayrulecolor{black}
    \cdashline{1-8}[1.5pt/4pt]

    GPT4RoI (arXiv-23)~\cite{zhang2023gpt4roi} & \textcolor{ForestGreen}{\usym{2713}} & \textcolor{ForestGreen}{\usym{2713}}~/~\textcolor{red}{\usym{2717}} & \textcolor{ForestGreen}{\usym{2713}}~/~\textcolor{red}{\usym{2717}} & \textcolor{ForestGreen}{\usym{2713}}~/~\textcolor{red}{\usym{2717}} & \textcolor{red}{\usym{2717}} & \textcolor{ForestGreen}{\usym{2713}} & \textcolor{ForestGreen}{\usym{2713}} \\

    \rowcolor{LGray} ASM (arXiv-23)~\cite{wang2023all} & \textcolor{ForestGreen}{\usym{2713}} & \textcolor{ForestGreen}{\usym{2713}}~/~\textcolor{red}{\usym{2717}} & \textcolor{red}{\usym{2717}}~/~\textcolor{red}{\usym{2717}} & \textcolor{ForestGreen}{\usym{2713}}~/~\textcolor{red}{\usym{2717}} & \textcolor{red}{\usym{2717}} & \textcolor{red}{\usym{2717}} & \textcolor{ForestGreen}{\usym{2713}} \\

    LISA (arXiv-23)~\cite{lai2023lisa} & \textcolor{ForestGreen}{\usym{2713}} & \textcolor{red}{\usym{2717}}~/~\textcolor{ForestGreen}{\usym{2713}} & \textcolor{red}{\usym{2717}}~/~\textcolor{red}{\usym{2717}} & \textcolor{red}{\usym{2717}}~/~\textcolor{ForestGreen}{\usym{2713}} & \textcolor{ForestGreen}{\usym{2713}} & \textcolor{red}{\usym{2717}} & ~\textcolor{ForestGreen}{\usym{2713}} \\

    \rowcolor{violet!10}GLaMM (ours) & \textcolor{ForestGreen}{\usym{2713}} & \textcolor{ForestGreen}{\usym{2713}}~/~\textcolor{ForestGreen}{\usym{2713}} & \textcolor{ForestGreen}{\usym{2713}}~/~\textcolor{ForestGreen}{\usym{2713}} & \textcolor{ForestGreen}{\usym{2713}}~/~\textcolor{ForestGreen}{\usym{2713}} & \textcolor{ForestGreen}{\usym{2713}} & \textcolor{ForestGreen}{\usym{2713}} & \textcolor{ForestGreen}{\usym{2713}} \\
        
        \bottomrule
    \end{tabular}
    }\vspace{-0.5em}
    \caption{\textbf{Comparison of recent Large Multimodal Models (LMMs)} emphasizing their capabilities for region-level understanding. The \textit{Input} denotes models that can process regions defined by users via bounding boxes, with \textit{Multi-Region} indicating models that can handle multiple such regions. The \textit{Output} represents models capable of delivering grounded responses. While some methods employ external vision modules for region understanding, others rely solely on the LMM, which may result in imprecise localization. However, a few integrate specialized vision modules and LMMs, as indicated by the \textit{Region Enc./Dec.}. The \textit{End-End Model} distinction separates models that leverage LMMs for region understanding from those employing external modules. \textit{Pixel-wise Grounding} highlights models that can respond with segmentation masks, and \textit{Multi-turn Conversation} represents models that can hold an interactive dialogue with the user. Among these, our proposed \textit{GLaMM} stands out by offering comprehensive region understanding, pixel-wise grounding in its responses, conversational capabilities, and an end-to-end training approach.}
    \label{tab:methods_comparison}
\vspace{-1em}
\end{table*}

%% file: sec/2_related.tex
\section{Related Work}
\label{sec:related}
\label{sec:related}
LMMs provide a 
versatile interface for a diverse array of tasks, encompassing language and vision. 
Prominent models such as  BLIP-2~\cite{li2023blip}, LLaVA~\cite{liu2023llava}, InstructBLIP~\cite{dai2023instructblip} and MiniGPT-4~\cite{zhu2023minigpt} first conduct image-text feature alignment followed by instruction tuning. 
Other representative works include Otter~\cite{li2023otter},  mPLUG-Owl~\cite{ye2023mplugowl}, LLaMa-Adapter~\cite{zhang2023llama}, Video-ChatGPT~\cite{maaz2023video}, InternGPT~\cite{liu2023interngpt}. 
However, these approaches lack region-specific understanding.

Recent works like Kosmos-2 \cite{peng2023kosmos}, Shikra \cite{chen2023shikra}, GPT4RoI \cite{zhang2023gpt4roi}, VisionLLM \cite{wang2023visionllm}, Ferret \cite{you2023ferret} and All-Seeing \cite{wang2023all} aim to allow region-specific conversation. Some methods \cite{peng2023kosmos,chen2023shikra,you2023ferret,wang2023all} input location bins and bounding boxes with image data for region-level understanding, relying on the LLM exclusively for interpreting these regions. GPT4RoI advances this by using spatial boxes and RoI-aligned features for input and training on region-text pairs. 
BuboGPT~\cite{zhao2023bubogpt} utilizes an off-the-shelf grounding model~\cite{liu2023grounding} and matches the groundings with the language response. In contrast, LISA \cite{lai2023lisa} utilizes embeddings from the vision language model and the SAM \cite{kirillov2023segment} decoder to generate output segmentation masks. However, LISA cannot comprehend specific image regions or handle multiple instances. 


To classify the LMM landscape, methods can be partitioned into four distinct categories (see Tab.~\ref{tab:methods_comparison} - separated via dotted lines). The first encompasses models effective in textual responses but lacking in region-specific capabilities~\cite{dai2023instructblip, gao2023llamaadapterv2, li2023otter, liu2023llava, yang2023mmreact, ye2023mplugowl, zhu2023minigpt}. In contrast, among models that handle region inputs or offer visual grounding, \textit{three} more categories emerge. The first of these incorporates external vision modules~\cite{liu2023interngpt, zhao2023bubogpt}, and the next relies exclusively on LMMs for region understanding~\cite{wang2023visionllm, pi2023detgpt, chen2023shikra, peng2023kosmos}. The last category combines specialized vision modules with LMMs, trained end-to-end for a comprehensive understanding of regions~\cite{zhang2023gpt4roi, wang2023all, lai2023lisa}.
Our approach belongs to the last category and distinctly offers pixel-level grounding together with multi-turn conversations and the flexibility to operate on both input images and specific regions.
Further, we provide  large-scale instance-level grounded visual understanding dataset that allows generalizability of  GLaMM  to multiple vision-language tasks.

%% file: sec/4_method.tex
\section{Method}
\label{sec:method}
Existing Large Multimodal Models (LMMs) either generate ungrounded text or are restricted by limitations such as single-object grounding, user-specified region inputs, or the lack of dense pixel-level object grounding (see Tab.~\ref{tab:methods_comparison}). 
Our Grounding LMM (GLaMM) aims to overcome these limitations by generating natural language responses seamlessly integrated with object segmentation masks. This enables a visually grounded human-machine conversation. 

\subsection{GLaMM Architecture}
\begin{figure*}[h!tp]
  \centering
    \includegraphics[width=0.98\linewidth]{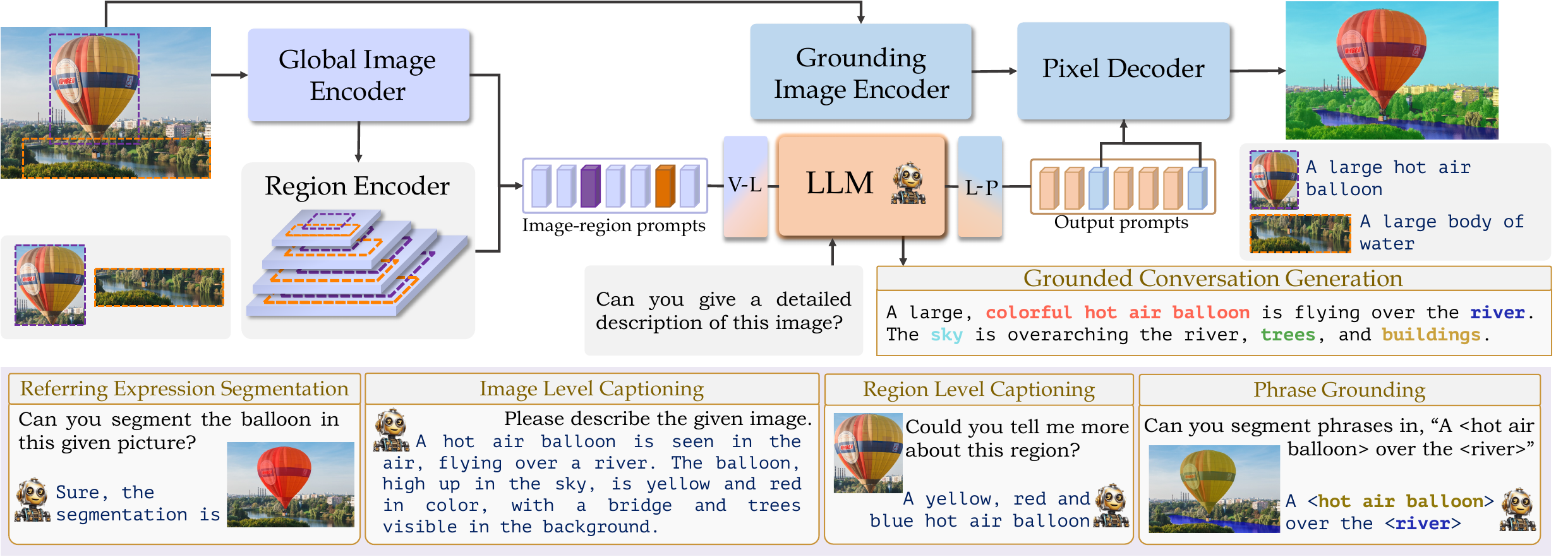}\vspace{-0.5em}
    \caption{\textbf{GLaMM's architecture.} The figure illustrates our model architecture, showcasing its ability to offer scene-level understanding, region-level interpretation, and pixel-level grounding. \textbf{Top:} The core components of GLaMM, including the global image encoder, region encoder, LLM, grounding image encoder, and pixel decoder, are cohesively tailored for vision-language tasks across different granularities. The vision-to-language (V-L) projection layer efficiently maps image features into the language domain, and the pixel decoder utilizes the language-to-prompt (L-P) projection layer, transforming text embeddings related to segmentation into the decoder space. A major feature of GLaMM is its ability to perform our newly introduced \textit{Grounded Conversation Generation (GCG)} task. This highlights the model's capability to anchor specific phrases to corresponding segmentation masks in the image. \textbf{Bottom}: The diverse downstream applications of GLaMM, including referring expression segmentation, region-level captioning, image-level captioning, and phrase grounding.}
    \label{fig:main_fig}
\vspace{-1em}
\end{figure*}

GLaMM consists of five core components: i) Global Image Encoder, ii) Region Encoder, iii) LLM, iv) Grounding Image Encoder, and v) Pixel Decoder. These components are cohesively designed to handle both textual and optional visual prompts (image level and region), allowing for interaction at multiple levels of granularity and generating grounded text responses (Fig.~\ref{fig:main_fig}). These blocks together enable scene-level, region-level, and pixel-level grounding, as explained next. 
Refer Appendix A.2 for training details.

\noindent \textbf{Scene-Level Understanding}: To achieve a holistic understanding of the scene, we employ ViT-H/14 CLIP~\cite{radford2021learning} as our \textit{global image encoder }$(\mathcal{I})$, in conjunction with a vicuna-based LLM $(\mathcal{L})$ and a vision-to-language (V-L) projection layer $(f)$. Specifically, given an image $x_{\text{img}}$ and a text instruction $x_t$, the image is first encoded into a feature vector $I_x = \mathcal{I}(x_{\text{img}}) \in \mathbb{R}^{D_v}$ and projected to language space $f({I_x}) \in \mathbb{R}^{D_t}$. The LLM then integrates both the projected image features and the text instruction to generate output $y_t$: 
\begin{align*}
y_t = \mathcal{L}\Bigl(f({I_x}), x_t \Bigr).
\end{align*}
This maps image features to language space, enabling GLaMM to offer holistic scene understanding, achieved through specific prompts like, 
``\texttt{The <image> provides an overview of the image. Could you please give me a detailed description of the image?}''  
The \texttt{<image>} token is replaced with 256 tokens from the CLIP global image encoder.

\noindent \textbf{Region-Level Understanding}: Building on the shortcomings of existing models that can handle only image-level visual inputs, and in alignment with recent work ~\cite{zhang2023gpt4roi}, the \textit{region encoder} $(\mathcal{R})$ extends the model's capability to interpret and interact with user-specified regions in an image. This component constructs a hierarchical feature pyramid from four selected CLIP global image encoder layers, followed by RoIAlign~\cite{he2017mask} to generate a 14x14 feature map. Combining these features yields a unified region-of-interest (RoI) representation. To facilitate region-targeted responses from GLaMM, we augment the existing vocabulary with a specialized token \texttt{<bbox>}. This is integrated into a prompt like,  ``\texttt{The <image> provides an overview of the image. Can you provide a detailed description of the region <bbox>?}''. Here the \texttt{<bbox>} token is replaced with the RoI extracted features.

For the region-level understanding, alongside the global image features $I_x$, we also take user-specified regions $ r $ as inputs, encoded as $ R_x = \mathcal{R}({I_x}, r) $, followed by projection to language space through the same V-L projection layer $f$ employed in scene-level understanding. We augment the text instruction $x_t$ by replacing \texttt{<bbox>} tokens with the corresponding region features to obtain $x_{t}' = [x_t \leftarrow f({R_x})]$. The LLM then generates the output $y_t$ as,
\begin{align*}
y_t = \mathcal{L}\Bigl(f({I_x}), x_{t}' \Bigr).
\end{align*}

\begin{figure*}[tp]
  \centering
    \includegraphics[width=0.98\linewidth]{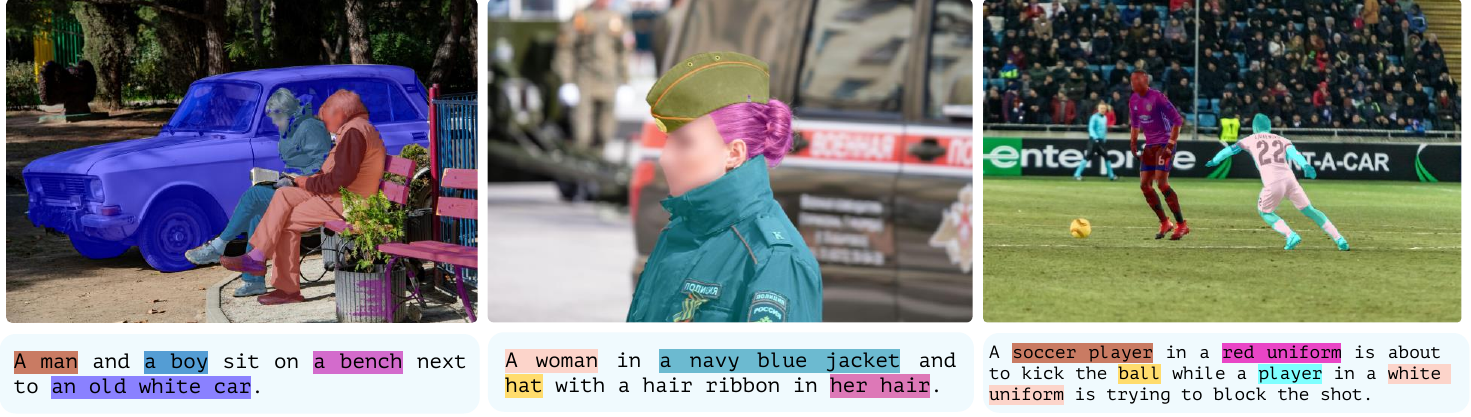}\vspace{-0.8em}
    \caption{\textbf{Qualitative results of GLaMM on grounded conversation generation (GCG).} Given user queries, the LMM generates textual responses and grounds objects, object parts, attributes, and phrases using pixel-level masks, showing its detailed understanding.}
    \label{fig:gcg}
\vspace{-1em}
\end{figure*}

\noindent \textbf{Pixel-Level Grounding}: Utilizing the \textit{grounding image encoder} denoted as $\mathcal{V}$ and the \textit{pixel decoder} represented as $\mathcal{P}$, GLaMM facilitates fine-grained pixel-level object grounding, allowing it to ground its responses visually. We instantiate $\mathcal{V}$ with a pretrained SAM encoder~\cite{kirillov2023segment} and design $\mathcal{P}$ based on a SAM decoder-like architecture. To activate the pixel-level grounding, our model's vocabulary is augmented with a specialized token, \texttt{<SEG>}. Prompts, such as ``\texttt{Please segment the `man in red' in the given image}," trigger the model to generate responses with corresponding \texttt{<SEG>} tokens. A \textit{language-to-prompt (L-P)} projection layer $(g)$ transforms the last-layer embeddings corresponding to \texttt{<SEG>} tokens $(l_{seg})$ into the decoder's feature space. Subsequently, $\mathcal{P}$ produces binary segmentation masks $M$,
\begin{align*}
M = \mathcal{P}\Bigl( g(l_{seg}), \mathcal{V}(x_{\text{img}}) \Bigr), \; s.t., M_i \in \{0,1\}.
\end{align*}
Using an end-to-end training approach, GLaMM excels in region understanding, pixel-level grounding, and conversational capabilities.
However, due to the lack of standard benchmarks for the novel setting of generating visually grounded detailed conversations, 
we introduce a novel task, \textit{Grounded Conversation Generation} (GCG), and a comprehensive evaluation protocol as explained next.

\subsection{Grounded Conversation Generation (GCG)}
\label{sec:gcg}
The objective of the GCG task is to construct image-level captions with specific phrases directly tied to corresponding segmentation masks in the image. For example, ``\texttt{<A man> and <a boy> sit on <a bench> next to <an old white car>.}'', shown in Fig.~\ref{fig:gcg} (left), features how each bracketed phrase (highlighted in the image) is anchored to a unique image segmentation mask. This creates a densely annotated caption that aligns textual descriptions with visual regions, enriching the image's contextual interpretation. 

\noindent \textbf{GCG Output Representation}:
A sample prompt for querying the model in this task is:  ``\texttt{Could you please give me a detailed description of the image? Please respond with interleaved segmentation masks for the corresponding parts of the answer.}'' The model generates a detailed caption along with interleaved segmentation masks, employing the format ``\texttt{<p>A man</p><SEG> and <p>a boy</p><SEG> sit on <p>a bench</p><SEG> next to <p>an old white car</p><SEG>.}'' We use special tokens, namely \texttt{<p>, </p>} and \texttt{<SEG>}, to delineate the start and end of each phrase and its corresponding region mask, respectively.

Our GranD dataset is meticulously constructed using a stage-wise annotation pipeline, capturing annotations that range from fine-grained specifics to high-level context. This enables the automatic generation of densely annotated captions well-suited for the GCG task, thereby significantly facilitating GLaMM's training for this task. Some qualitative results of our model on the GCG task are shown in Fig.~\ref{fig:gcg}.

\noindent \textbf{Evaluation Criteria}:
We introduce a benchmarking suite for GCG, with a validation set of 2.5K images and a test set of 5K images. Four key aspects are evaluated: i) generated dense caption quality, ii) mask-to-phrase correspondence accuracy, iii) generated mask quality, and iv) region-specific grounding ability. Metrics include METEOR and CIDEr for captions, class-agnostic mask AP for grounding, mask IoU for segmentation, and mask recall for region-specific grounding 
(refer to Appendix~A.1 for details).

Having delineated the architecture of GLaMM and the intricacies of the GCG task, it becomes imperative to address the scarcity of large-scale annotated data for region-level understanding. We next focus on devising a new, densely annotated dataset to optimize the model's performance and overcome this data limitation.

\begin{figure*}[!th]
\centering
{\includegraphics[width=0.98\textwidth]{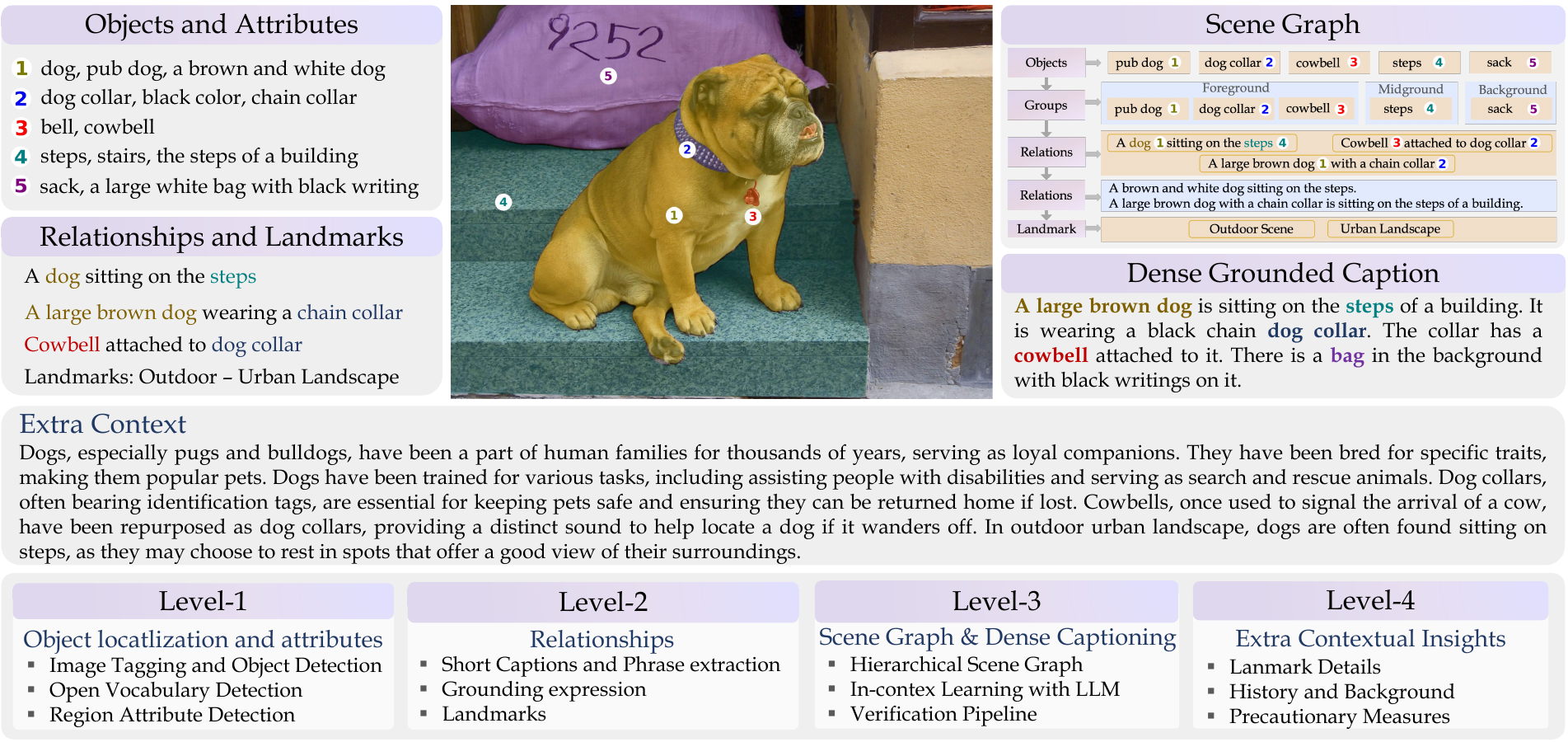}}\vspace{-0.5em} \caption{\small \textbf{Automatic Annotation Pipeline of the Grounding-anything Dataset (GranD).} Comprising four levels, this pipeline plays a pivotal role in generating GranD's 7.5M unique concepts grounded in 810M regions. level-1 details objects and attributes, level-2 includes short captions and relational markers, level-3 builds a scene graph, hierarchically organizing information from earlier levels to facilitate LLM for grounded dense captions, level-4 provides additional historical and societal context for a richer visual understanding.}
\label{fig:dataset_pipeline}
\vspace{-1em}
\end{figure*}


%% file: sec/3_dataset.tex
\section{Data Annotation Pipeline}
\label{sec:dataset}
We introduce our automated annotation pipeline used to create the Grounding-anything Dataset (GranD). GranD is a comprehensive, multi-purpose image-text dataset offering a range of contextual information, from fine-grained to high-level details. It aims to overcome challenges in image understanding and dense pixel-level grounding, thereby expanding capabilities of visual instruction tuning in LMMs.

The pipeline contains four distinct levels
 (see Fig.~\ref{fig:dataset_pipeline}).

\noindent
\textbf{i)} \textit{Level-1} focuses on object localization and provides semantic labels, segmentation masks, attributes, and depth information.
\textbf{ii)} \textit{Level-2} defines relationships between detected objects. 
\textbf{iii)} \textit{Level-3} organizes information from the first two levels into a hierarchical scene graph, used to generate dense captions using LLM with in-context examples.
\textbf{iv)} \textit{Level-4} offers enriched contextual information for a deeper understanding of the scene, going beyond what's observed (e.g., historical information of a landmark).  Please refer to
Appendix A.4 
for pipeline implementation details.

\subsection{Object Localization and Attributes (Level-1)}
In level-1, the focus is on detailed object identification within images. First, object-bounding boxes are identified using multiple SoTA object detection models. Class-agnostic NMS is applied to each model to filter out false positives. After this step, bounding boxes from different models are compared using IoU, with a bounding box retained as an object only if detected by at least two other detection models. We also generate attributes for each filtered object using region-based vision-language models and incorporate depth information to contextualize each object's relative position within the scene.

\subsection{Relationships and Landmarks (Level-2)}
In level-2, multiple short textual descriptions of the overall scene are generated. Phrases extracted from these descriptions are grounded to specific objects in level-1 to form relationships. These relationships articulate connections between multiple objects or define an object's role within the scene. Further, each scene is assigned a landmark category that includes a primary and a more specific sub-category 
(see Tab.~7 in Appendix~A.4.1).

\subsection{Scene Graph and Dense Captioning (Level-3)}
In level-3, object attributes and labels from level-1 are combined with the relationships and phrases obtained from level-2 to form a hierarchical scene graph. This structured data serves as a query for LLM to generate dense image captions. To provide additional context, depth values and bounding box coordinates are used to assign each object to specific spatial layers within the scene, such as \emph{immediate foreground}, \emph{foreground}, \emph{midground}, or \emph{background}. Additionally, short scene-level captions are incorporated into the scene graph to enhance LLMs' contextual understanding.

\noindent
\textbf{Dense Captioning Verification:} To enhance the fidelity of the LLM-generated dense captions, we implement an automatic verification pipeline using chain-of-thoughts prompting. This pipeline produces a checklist of objects derived from the generated dense caption expected to be present in the image. The associated caption is flagged as inaccurate if any object specified in the checklist is absent from the scene graph. Such captions are then regenerated, incorporating feedback from the initial assessment.

\subsection{Extra Contextual Insights (Level-4)}
Level-4 builds on the scene graph from level-3 to obtain a more detailed visual understanding. 
we query LLM to extract extended contextual insights beyond basic object identification and relationships, including details about the landmarks, historical context, guidelines for interacting with the scene, and even predictive elements about future events. 
To facilitate this, we prompt LLM with in-context examples.

Utilizing our automated annotation pipeline, we annotate a corpus of 11M SAM images~\cite{kirillov2023segment}, which are inherently diverse, high-resolution, and privacy-compliant. The resulting dataset comprises 810M regions, each associated with a segmentation mask, and includes 7.5M unique concepts. Further, the dataset features 84M referring expressions, 22M grounded short captions, and 11M densely grounded captions. To our knowledge, this is the first dataset of this scale generated entirely through an automated annotation pipeline (see Tab.~\ref{tab:dataset-comparisons} for details and 
Fig.~15 in Appendix for dataset sample visualizations).

\begin{table}[!t]
        \centering
        \small
\resizebox{1\columnwidth}{!}{
	\begin{tabular}{lccccc}
  \toprule
            \textbf{Dataset} & \textbf{Images} & \textbf{Regions} & \textbf{Concepts} & \textbf{Tokens} & \textbf{Captions}${^\dagger}$ \\
            \midrule
		COCO~\cite{lin2014microsoft}                     & 0.1M  & 0.9M   & 80     &  -     & - \\
		LVIS~\cite{gupta2019lvis}                         & 0.1M  & 1.5M   & 1,203  &  -     & - \\
		Objects365~\cite{shao2019objects365}           & 0.6M  & 10.1M  & 365    &   -    & - \\
		Open Images~\cite{kuznetsova2020open}       & 1.5M  & 14.8M  & 600    &  -     & - \\
		BigDetection~\cite{cai2022bigdetection}        & 3.5M  & 36.0M  & 600    &  -     & - \\
		V3Det~\cite{wang2023v3det}                    & 0.2M  & 1.5M   & 13,029 & -      & - \\
		VG~\cite{krishna2017visual}      & 0.1M  & 0.3M   & 18,136 &  51.2M   & - \\
        SA-1B~\cite{kirillov2023segment}                   & 11M     & 1.1B       & -     & -     & - \\
        \textcolor{gray}{AS-1B~\cite{wang2023all}}                                      & \textcolor{gray}{11M}    & \textcolor{gray}{1.2B}       & \textcolor{gray}{3.5M}     & \textcolor{gray}{132.2B}   & - \\
       \rowcolor{violet!10} GranD (Ours)                                     & 11M    & 810M       & 7.5M     & 5.0B     & 33M \\
  \bottomrule
	\end{tabular}}
        \setlength{\abovecaptionskip}{0.1cm}
        \caption{\textbf{GranD versus existing datasets.} GranD uniquely provides three $^{\dagger}$grounded captions per image with segmentation masks for every region. AS-1B is shaded to denote its concurrent, non-public status at the time of this publication.}
        \label{tab:dataset-comparisons}
\vspace{-1.5em}
\end{table}

\begin{SCtable*}[][!h]
    \tabstyle{5pt}
    \scalebox{1}{
    
    \begin{tabular}{lccccc|ccccc}
    \toprule
    \multirow{2}{*}{Model} & \multicolumn{5}{c|}{Validation Set} &  \multicolumn{5}{c}{Test Set} \\ 
                           & M & C & AP50 & mIoU & Recall & M & C & AP50 & mIoU & Recall\\ \midrule
    BuboGPT~\cite{zhao2023bubogpt}& \textbf{17.2}   & 3.6  & 19.1 & 54.0 & 29.4 & \textbf{17.1}  & 3.5  & 17.3 & 54.1 & 27.0 \\
    Kosmos-2~\cite{peng2023kosmos}& 16.1   & 27.6  & 17.1 & 55.6 & 28.3 & 15.8   & 27.2  & 17.2 & 56.8 & 29.0 \\
    LISA*~\cite{lai2023lisa}& 13.0   & 33.9  & 25.2 & 62.0 & 36.3 & 12.9   & 32.2  & 24.8 & 61.7 & 35.5 \\
    \rowcolor{violet!10} GLaMM$\dagger$              & 15.2   & 43.1  & 28.9 & 65.8 & 39.6 & 14.6 & 37.9  & 27.2 & 64.6 & 38.0 \\ 
    
    \rowcolor{violet!10} GLaMM               & 16.2   & \textbf{47.2}  & \textbf{30.8} & \textbf{66.3} & \textbf{41.8} & 15.8 & \textbf{43.5}  & \textbf{29.2} & \textbf{65.6} & \textbf{40.8} \\ \hline
    
    \end{tabular}
}
\caption{\textbf{Performance on GCG Task}: Metrics include METEOR (M), CIDEr (C), AP50, mIoU, and Mask Recall. LISA* denotes LISA adapted for GCG. GLaMM$\dagger$ denotes training excluding 1K human annotated images. GLaMM shows better performance.}
\label{results_table1}
\end{SCtable*}

\begin{SCtable*}[][!h]
    \tabstyle{6pt}
    \scalebox{1}{
\begin{tabular}{lccccccccccc}
\toprule
\textbf{Method} && \multicolumn{3}{c}{\textbf{refCOCO}} && \multicolumn{3}{c}{\textbf{refCOCO+}} && \multicolumn{2}{c}{\textbf{refCOCOg}} \\
\cmidrule(lr){3-5} \cmidrule(lr){7-9} \cmidrule(lr){11-12}
&& val & testA & testB && val & testA & testB && val(U) & test(U) \\
\midrule
CRIS~\cite{wang2022cris}&& 70.5 & 73.2 & 66.1 && 65.3 & 68.1 & 53.7 && 59.9 & 60.4 \\
LAVT~\cite{yang2022lavt}&& 72.7 & 75.8 & 68.8 && 62.1 & 68.4 & 55.1 && 61.2 & 62.1 \\
GRES~\cite{liu2023gres}&& 73.8 & 76.5 & 70.2 && 66.0 & 71.0 & 57.7 && 65.0 & 66.0 \\
X-Decoder~\cite{zou2023generalized}&& - & - & - && - & - & - && 64.6 & - \\
SEEM~\cite{zou2023seem}&& - & - & - && - & - & - && 65.7 & - \\
LISA-7B~\cite{lai2023lisa}&& 74.9 & 79.1 & 72.3 && 65.1 & 70.8 & 58.1 && 67.9 & 70.6 \\
\rowcolor{violet!10} GLaMM && \textbf{79.5} & \textbf{83.2} & \textbf{76.9} && \textbf{72.6} & \textbf{78.7} & \textbf{64.6} && \textbf{74.2} & \textbf{74.9} \\
\bottomrule
\end{tabular}
}
\caption{\textbf{Qualitative Assessment of GLaMM in Referring-Expression Segmentation}: Performance across refCOCO, refCOCO+, and refCOCOg in generating accurate segmentation masks based on text-based referring expressions surpasses that of closely related work, including LISA which is specifically designed for this task.}
\label{results_table3}
\end{SCtable*}

\subsection{Building GranD$_f$ for GCG}
\label{sec:data_gcg}
Motivated by the need for higher-quality data in fine-tuning stage, we introduce GranD$_f$. 
It contains 214K image-grounded text pairs with 2.5K validation and 5K test samples. 
GranD$_f$ comprises two primary components: one subset is manually annotated, and the other subset is derived by re-purposing existing open-source datasets.

We extend open-source datasets—namely Flickr-30K~\cite{plummer2015flickr30k}, RefCOCOg~\cite{kazemzadeh2014referitgame}, and PSG~\cite{yang2022psg} by generating compatible GCG annotations. For RefCOCOg, we use the dataset's referring expressions and their connected masks. These expressions offer concise descriptions of distinct objects in the image. With the aid of GPT-4, we seamlessly blend these referring expressions with contextual information from COCO captions, crafting detailed yet accurate grounded captions while preserving the original referring expressions. This ensures zero error in matching phrases with their corresponding segmentation masks. This technique yields approximately 24K GCG samples. For PSG, we leverage the dataset's triplet structures, which describe relations between two objects in a scene. These triplets are integrated with COCO captions using GPT-4, resulting in densely annotated captions that can be mapped to segmentation masks. This gives us around 31K additional GCG samples. For Flickr-30K, we use the 158K Flickr captions and their referring expressions alongside associated bounding boxes. These boxes are then accurately segmented using HQ-SAM~\cite{ke2023segment}. 

In addition, we contribute a minor, high-quality manual annotation set to benchmark the GCG task. Using GranD's automatic annotations as a base, annotators refine referring expressions to match SAM GT masks, yielding around 1000 focused samples for training and 1000 for evaluation (refer to 
Appendix D and Fig.~14 in Appendix
for designed prompts and dataset visualizations).

%% file: sec/5_experiments.tex
\section{Experiments}
\label{sec:exp}
We perform quantitative evaluations of GLaMM on six benchmarks: i) Grounded Conversation Generation (GCG), ii) referring-expression segmentation, iii) region-level captioning, iv) image-level captioning, v) conversational-style question answering and vi) phrase grounding. We present the first four benchmarks next, and the remaining are discussed in 
Appendix~B.
\setlength{\belowcaptionskip}{-8pt}
\noindent
\textbf{Grounded Conversation Generation (GCG).}
We pretrain GLaMM on GranD dataset followed by fine-tuning on the GranD$_f$ dataset. The results are presented in Tab.~\ref{results_table1} on both validation and test splits of the GranD$_f$ dataset (refer to Sec.~\ref{sec:gcg} and Sec.~\ref{sec:data_gcg} for details).
GLaMM shows improved performance compared to baseline methods. Pretrained models for BuboGPT and Kosmos-2 are sourced from official releases, and LISA is adapted and trained on the GranD$_f$ dataset for the GCG task. GLaMM$\dagger$ denotes the variant trained on GranD$_f$ dataset excluding the 1000 human-annotated images.
Qualitative results are shown in Fig.~\ref{fig:gcg} and supplementary 
Fig.~7.

\begin{figure*}[!t]
  \centering
    \includegraphics[width=0.99\linewidth]{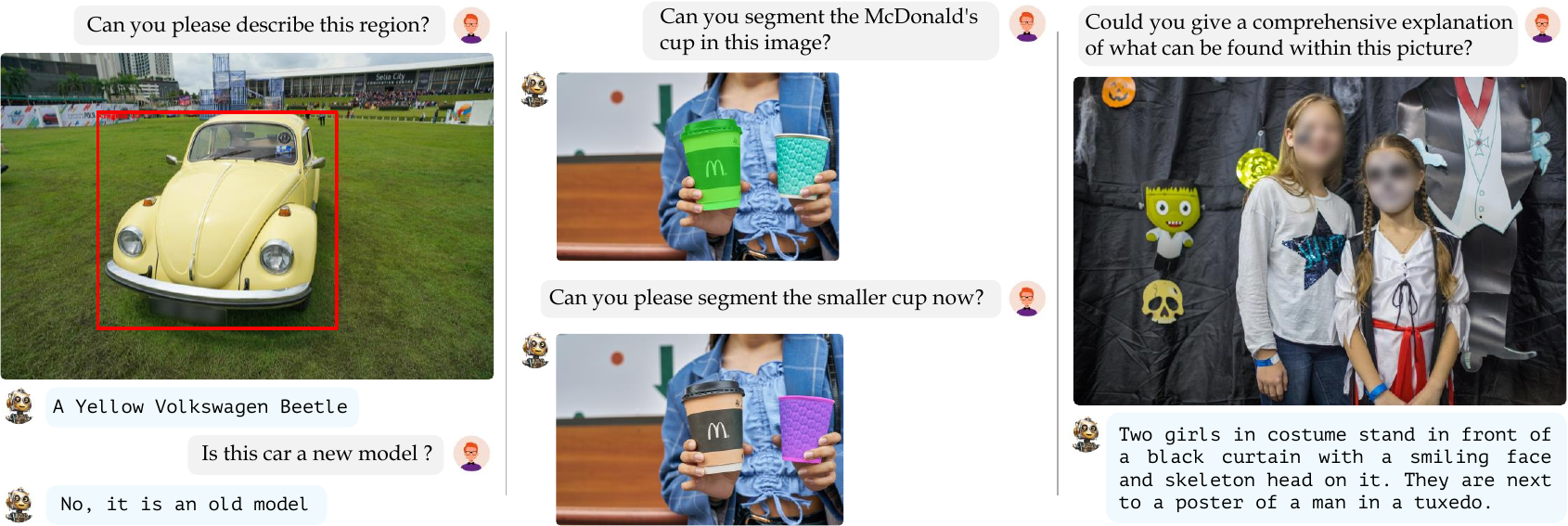}
    \caption{\textbf{Qualitative results of GLaMM's performance across downstream tasks.} The figure showcases examples from three tasks: region-level understanding (left), referring-expression segmentation (center), and image-level captioning (right), demonstrating its capabilities in offering in-depth region understanding, pixel-level groundings, and conversational abilities through an end-to-end training approach.}
    \label{fig:downstream}
\end{figure*}

\noindent
\textbf{Referring Expression Segmentation.} 
In this task, the model processes an image and a text-based referring expression to output a segmentation mask. The prompt used is, ``\texttt{Please segment the \texttt{<referring expression>} in the image.}"
The model responds with ``\texttt{Sure, it is \texttt{<SEG>}.}", where the \texttt{<SEG>} token is decoded to obtain the mask.
We achieve better results over recent works like LISA on the refCOCO, refCOCO+, and refCOCOg validation and test sets in Tab.~\ref{results_table3}.
This demonstrates the efficacy of our GranD dataset, offering the model extensive concept vocabulary during pre-training (refer to Fig.~\ref{fig:downstream} (middle) and supplementary 
Fig.~8  
for qualitative results).

\begin{table}[!t]
\centering
\resizebox{0.98\columnwidth}{!}{
\begin{tabular}{lcccc}
\toprule
\textbf{Model} & \multicolumn{2}{c}{\textbf{refCOCOg}} & \multicolumn{2}{c}{\textbf{Visual Genome}} \\ 
\cmidrule(lr){2-3} \cmidrule(lr){4-5} & METEOR & CIDEr & METEOR & CIDEr \\ 
\midrule
GRIT \cite{wu2022grit}& 15.2 & 71.6 & 17.1 & 142 \\
Kosmos-2 \cite{peng2023kosmos}& 14.1 & 62.3 & - & - \\
GPT4RoI \cite{zhang2023gpt4roi}& - & - & 17.4 & 145.2 \\
\rowcolor{violet!10} GLaMM (ZS) &\textbf{15.7} & \textbf{104.0} & \textbf{17.0} & \textbf{127.0} \\
\rowcolor{violet!10} GLaMM (FT) &\textbf{16.2} & \textbf{106.0} & \textbf{19.7} & \textbf{180.5} \\
\bottomrule
\end{tabular}
}
\caption{\textbf{Performance of GLaMM in Region-Level Captioning}: Metrics include METEOR and CIDEr scores, assessed on Visual Genome and refCOCOg Datasets, exhibiting competitive results.}
\label{results_table2}
\end{table}

\begin{table}[!th]
\centering
\resizebox{0.98\columnwidth}{!}{
\begin{tabular}{lcccc}
\toprule
\textbf{Model} & \multicolumn{2}{c}{\textbf{NoCap}} & \multicolumn{2}{c}{\textbf{Flickr30k}} \\
\midrule
& CIDEr & SPICE & CIDEr & SPICE \\
\midrule
VinVLM~\cite{zhang2021vinvl}          & 95.5 & 13.5 & - & - \\
LEMON~\cite{hu2022scaling}           & 106.8 & 14.1 & - & - \\
SimVLM~\cite{wang2021simvlm}         & 110.3 & 14.5 & - & - \\
CoCa~\cite{yu2022coca}            & 120.6 & 15.5 & - & - \\
BLIP~\cite{li2022blip}            & 113.2 & 14.7 & - & - \\
BLIP-2~\cite{li2023blip}          & 121.6 & 15.8 & - & - \\
InstructBLIP~\cite{dai2023instructblip}    & \textbf{123.1} & - & 82.8 & - \\
Shikra-13B~\cite{chen2023shikra}       & - & - & 73.9 & -\\
Kosmos-1~\cite{kosmos1}       & - & - & 67.1 & 14.5 \\
Kosmos-2~\cite{peng2023kosmos}        & - & - & 66.7 & - \\
\rowcolor{violet!10}
GLaMM            & 106.8 & \textbf{15.8} & \textbf{95.3} &\textbf{ 18.8} \\
\bottomrule
\end{tabular}
}
\caption{\textbf{Performance of GLaMM in Zero-Shot Image Captioning}: Assessed on Flickr30k and NoCap datasets, showing favorable results compared to recent models in the field.}
\label{results_table4}
\vspace{-1.5em}
\end{table}

\noindent
\textbf{Region Level Captioning.}
In this task, models generate region-specific captions given an image, a user-specified region via a bounding box and related text.
We utilize a prompt like, ``\texttt{Can you provide a detailed description of the region \texttt{<bbox>}?}'', to instruct the model for this task, where the special token \texttt{<bbox>} is replaced with the actual region representations. 
We evaluate GLaMM on Visual Genome and refCOCOg, using METEOR and CIDEr metrics with results presented in Tab.~\ref{results_table2}. GLaMM shows improved results over GRiT and GPT4RoI after fine-tuning and demonstrates robust zero-shot performance, highlighting the significance of GranD's region-text pairs (refer to Fig.\ref{fig:downstream} (left) and supplementary 
Fig.~9 for qualitative results).

\noindent
\textbf{Image Level Captioning.}
For this task, GLaMM responds to queries like, ``\texttt{Could you please give me a detailed description of the image?}" with a textual description. We evaluate GLaMM's zero-shot performance on Flickr30k~\cite{plummer2015flickr30k} and NoCap~\cite{agrawal2019nocaps} datasets, with Tab.~\ref{results_table4} showing its favorable performance against recent image captioning models and other LMMs (refer to Fig.~\ref{fig:downstream} (right) and supplementary 
Fig.~10 for qualitative results). 
\setlength{\belowcaptionskip}{0pt}


%% file: sec/conclusion.tex
\section{Conclusion}
\label{sec:conclusion}
We introduce GLaMM, the first model capable of generating natural language responses intertwined with object segmentation masks, allowing for enhanced multimodal user interactions. 
Recognizing the lack of standardized benchmarks for visually grounded conversations, we introduce the novel task of Grounded Conversation Generation and establish a comprehensive evaluation protocol. 
To facilitate research and model development, we create the Grounding-anything Dataset (GranD), a large-scale, densely annotated dataset with 7.5 million unique concepts grounded in 810 million regions. 
Our automated annotation pipeline ensures the reliability and scalability of this dataset used for our model.
In addition to these contributions, we curated a dataset specifically tailored for the GCG task (GranD$_f$) by leveraging existing open-source datasets, establishing a high-quality fine-tuning dataset to develop visually grounded conversations. 
Our model performs well on downstream tasks besides GCG, including region and image captioning, referring segmentation, and vision-language conversations.

%% file: sec/X_suppl.tex
\clearpage
\setcounter{page}{1}
\maketitlesupplementary
\appendix

\noindent We provide supplementary material for a deeper understanding and more analysis related to the main paper, arranged as follows:
\begin{enumerate}
    \item Additional implementation details (Appendix~\ref{sup:implementation_details})
    \item Additional downstream tasks (Appendix~\ref{sup:ds_tasks}
    \item Additional qualitative results (Appendix~\ref{sup:additional_qualitative_results})
    \item Dataset visualizations (Appendix~\ref{sup:dataset_viz})
    \item Limitations and future work (Appendix~\ref{sup:limitations})
    \item Ethics and societal impact (Appendix~\ref{sup:ethics})
\end{enumerate}

\section{Additional Implementation Details}
\label{sup:implementation_details}

\subsection{Evaluation Metrics}
\label{sup:eval_metrics}
\noindent \textbf{Mask Recall}:
To quantify region-specific grounding, we propose a `mask recall' metric, utilizing a two-tiered validation approach. Initially, predicted masks are mapped to ground-truth masks via a one-to-one set assignment, followed by IoU computation for these pairs. Pairs surpassing a 0.5 IoU threshold proceed to a textual similarity assessment using BERT. A pair is considered a true positive (TP) only if both IoU and BERT similarity exceed their 0.5 thresholds; otherwise, it is classified as a false positive (FP). The mask recall is subsequently calculated using the standard formula, normalizing the number of TPs by the total ground-truth mask count.

\subsection{Model Architecture and Training}
\label{sup:app_training}
In all of our experiments, we use Vicuna LLM~\cite{vicuna} with 7B parameters. The design of region encoder is motivated from GPT4RoI~\cite{zhang2023gpt4roi} and grounding image encoder and pixel decoder are inspired from LISA~\cite{lai2023lisa}. The V-L and L-P layers are implemented using 2 layer MLP with GELU activation as in LLaVA-v1.5~\cite{liu2023improvedllava}. We use PyTorch to implement our GLaMM and use Deepspeed zero-2 optimization during training. 

Specifically, our model is trained using two types of losses: auto-regressive cross-entropy loss for text generation and a linear combination of per-pixel binary cross-entropy loss and DICE loss for segmentation. During training, the global image encoder and grounding image encoder are kept frozen and the region encoder, projection layers (V-L and L-P) and the pixel decoder are fully finetuned, while the LLM is LORA finetuned with $\alpha=8$. Our codes and pretrained models will be publicly released.

\subsubsection{Pretraining on GranD}
During pretraining GLaMM is trained on GranD dataset for referring expression segmentation, region-level captioning, image-level captioning and grounded conversation generation (GCG) tasks simultaneously. We use a batch size of 160 and train for a total of 35K iterations during pretraining. We use LORA-8 for efficiently adapting the LLM and initialize the pretraining from GPT4RoI~\cite{zhang2023gpt4roi} for faster convergence. In the experiment tables in Section.~\ref{sec:exp}, we refer to this model as GLaMM (ZS) which is obtained after pretraining on GranD.

\subsection{Finetuning on Downstream Tasks}
We finetune GLaMM on multiple downstream tasks including GCG, referring expression segmentation, region-level captioning and image-level captioning. For GCG, we finetune our model on GranD$_f$ dataset. A batch size of 160 is used and the model is trained for 5K iterations in total. It is worth noting that GranD$_f$ dataset is a combination of multiple open-source datasets that we repurposed for GCG task using GPT4~\cite{openai2023gpt4}. Please refer to Appendix.~\ref{sup:dataset_viz} for the prompts designed to query GPT4 for constructing GranD$_f$ dataset, along with the dataset visualizations.

For referring expressions segmentation, we finetune GLaMM on refCOCO, refCOCO+ and refCOCOg datasets. We represent this model as GLaMM (FT) in Tab.~\ref{results_table3}. Similarly, for region-level captioning, GLaMM (FT) is finetuned on refCOCOg and Visual Genome datasets. For image-level captioning, we fine tune GLaMM on LLaVA-Instruct-150K~\cite{liu2023llava} dataset. For LLaVA-bench, the model is finetuned on LLaVA-Instruct-80K~\cite{liu2023llava} instruction set. We use eight NVIDIA A100-40GB GPUs in all of our pretraining and finetuning experiments.

\begin{figure*}[t]
  \centering
    \includegraphics[width=\linewidth]{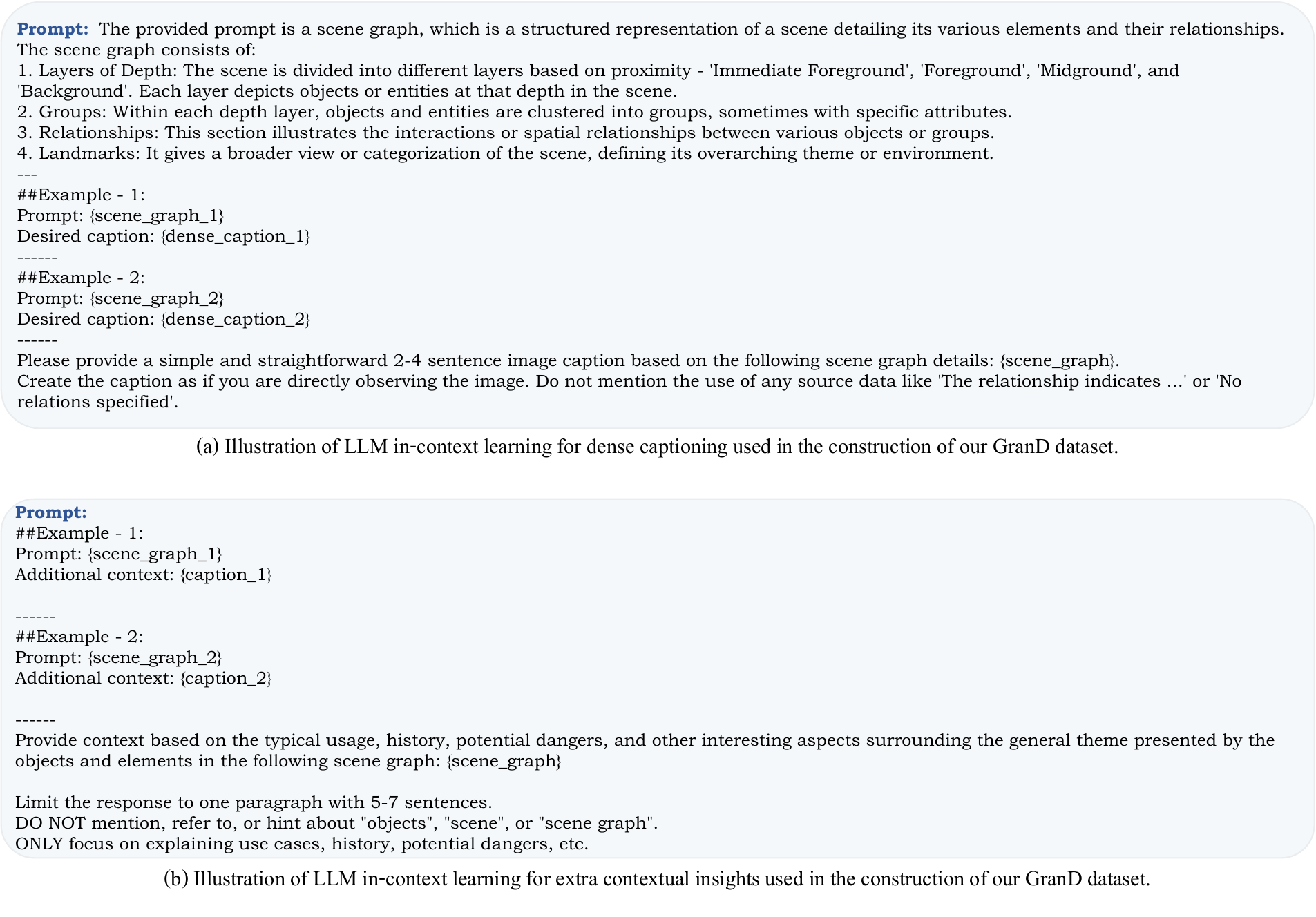}
    \caption{\textbf{Prompts used to construct GranD dataset.} The figure shows the prompts used to query Vicuna~\cite{vicuna} to generate dense captions and the extra context in our automated training pipeline. We provide in-context examples to guide the LLM.}
    \label{fig:dataset_promp}
\end{figure*}

\subsection{Automated Dataset Annotation Pipeline}
\label{sup:annotation_pipeline}
Our automated annotation pipeline incorporates diverse state-of-the-art models at various levels. For Level-1, we use Tag2Text~\cite{huang2023tag2text} and RAM~\cite{zhang2023recognize} for image tagging, Co-DETR~\cite{codetr}, EVAv02~\cite{fang2023eva}, OWL-ViT~\cite{minderer2022simple}, and POMP~\cite{ren2023prompt} for object localization, GRiT~\cite{wu2022grit} and GPT4RoI~\cite{zhang2023gpt4roi} for attribute generation, and MiDAS~\cite{midas} for depth estimation. Level-2 leverages BLIP-2~\cite{li2023blip} and LLaVA-v1.5~\cite{liu2023llava, liu2023improvedllava} for scene descriptions and landmark categorization, SpaCy~\cite{spacy} for phrase extraction, and MDETR~\cite{kamath2021mdetr} for phrase grounding. For both Level-3 and Level-4, we use Vicuna-v1.5~\cite{vicuna} with 13B parameters, supplemented with in-context examples. Please refer to Appendix~\ref{sup:annotation_pipeline} for further details on implementation and LLM prompts used across different pipeline levels.

We design a fully automated dataset annotation pipeline using multiple hierarchical levels in the visual domain to construct GranD dataset. The segmentation masks for most of the regions are obtained from SAM~\cite{kirillov2023segment} annotations by comparing our detected labeled regions with SAM-provided class-agnostic regions. For the remaining regions that do not match with any of the SAM regions, we run SAM model with a bounding box query to obtain masks.

Our automated annotation pipeline utilizes only open-source models and incorporates a feedback loop using the chain of thoughts prompting via LLM. As it does not require feedback from the human in the loop, it can be scaled to generate dense noisy labels for a larger number of images, which can then be used to pretrain a larger LMM. Given the availability of enough compute power, this could be a step towards building a larger generic large multi-modal model. We will release our GranD dataset along with the implementation of our automated dataset annotation pipeline for further research. Below we present the LLM prompts we use at different levels of our automated dataset annotation pipeline.

\subsubsection{LLM Prompts and In-context Learning}
\label{sup:llm_prompts}
\noindent \textbf{Landmark categorization}: We use LLaVA-v1.5-13B~\cite{liu2023improvedllava} model to assign landmark categories to each image. Please refer to Tab.~\ref{tab:landmarks} for primary and fine categories used.

\begin{table}[h]
\centering
\setlength{\tabcolsep}{1mm}{
\resizebox{0.99\columnwidth}{!}{
\begin{tabular}{p{2.5cm}p{6.5cm}} 
\toprule
\rowcolor{violet!10} Main category             & Fine Category \\
\midrule
Indoor scene              & Living space, Work space, Public space, Industrial space \\
Outdoor scene             & Urban landscape, Rural landscape, Natural landscape \\
Transportation scene      & Road, Airport, Train station, Port and harbor \\
Sports and recreation scene & Sporting venue, Recreational area, Gym and fitness center \\
\bottomrule
\end{tabular}
}}
\caption{Summary of landmark categories and their corresponding fine-grained categories. We use LLaVA-v1.5~\cite{liu2023improvedllava} for assigning landmark categories to images.}
\label{tab:landmarks}
\end{table}

\noindent \textbf{Dense Captioning}: We arrange objects, attributes and relationships hierarchically to construct a visual scene graph, that is used to query Vicuna-v1.5-13B~\cite{vicuna} model along with in-context examples to generate dense captions. The designed prompt is shown in Fig.~\ref{fig:dataset_promp} (a).


\noindent \textbf{Extra Context}: We query Vicuna-v1.5-13B model to generate additional context about the visual scene. The prompt designed for this purpose is shown in Fig.~\ref{fig:dataset_promp} (b).



\section{Additional Downstream Tasks}
\label{sup:ds_tasks}

\subsection{Phrase Grounding}
In order to adapt the GLaMM model for phrase grounding, we repurpose the GCG dataset to suit this particular task. Specifically, the answers in the GCG dataset are now used as questions, and the parts of the captions containing groundings are regarded as phrases. The model is subsequently trained to locate pixel-level groundings for these phrases, which are enclosed within <p> and </p> tokens. The results of this adaptation are shown in the following figure.

\begin{figure}[!th]
  \centering
    \includegraphics[width=0.99\linewidth]{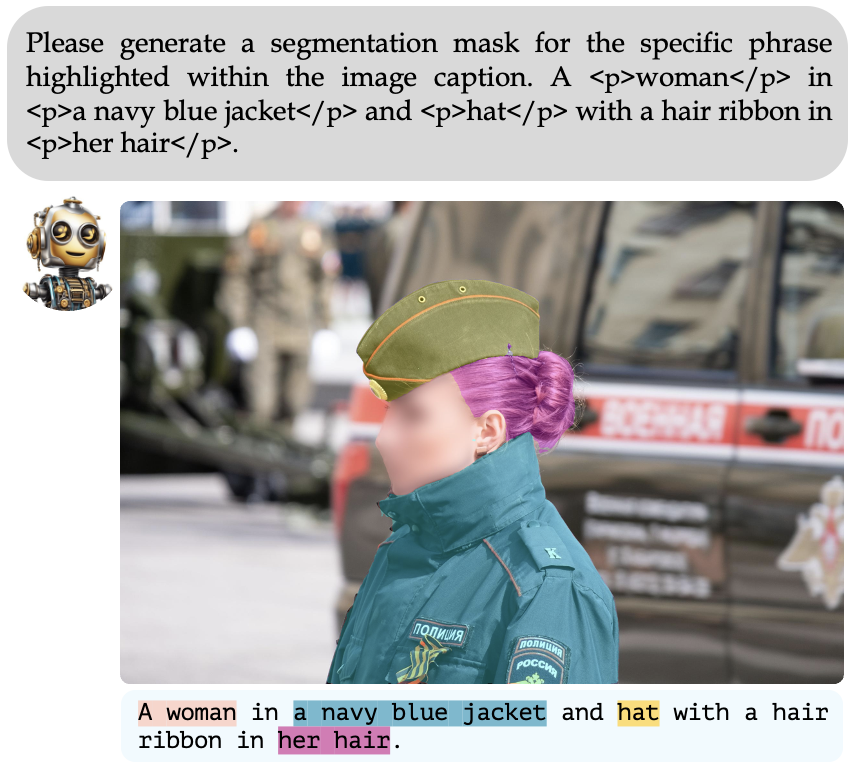}
    \label{fig:app_gcg2}
\end{figure}

\subsection{Conversational Style Question Answering}
\begin{figure*}[!th]
  \centering
    \includegraphics[width=\linewidth]{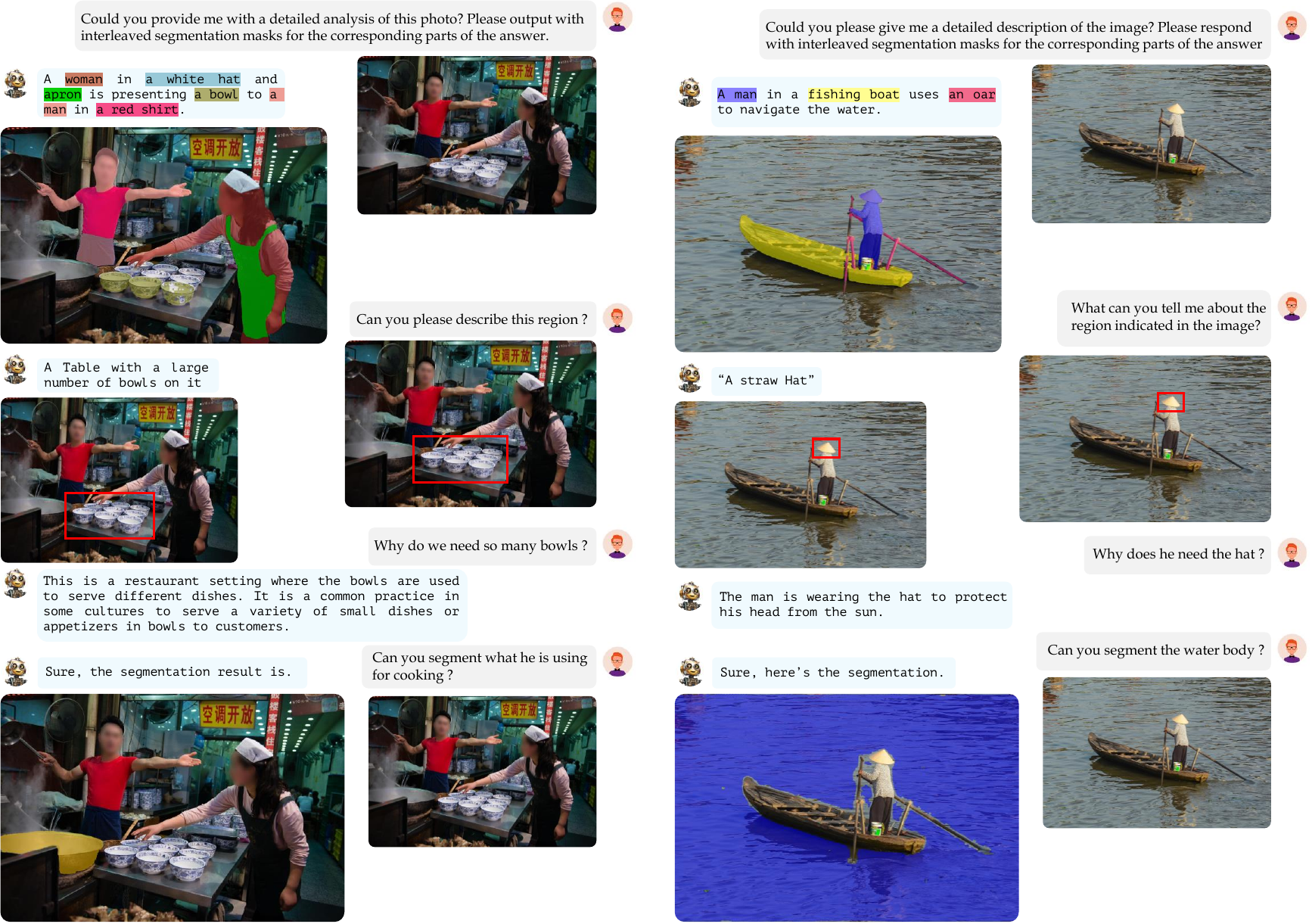}
    \caption{\textbf{Multimodal conversational interactions facilitated by GLaMM.} The figure showcases GLaMM engaging in multi-turn dialogues, providing detailed descriptions, addressing region-specific inquiries, and presenting grounded conversations. This effectively highlights its adaptability in intricate visual-language interactions and robustly retaining reasoning capabilities inherent to LLMs.}
    \label{fig:conv_main}
\end{figure*}

\begin{figure*}[!t]
  \centering
    \includegraphics[width=0.85\linewidth]{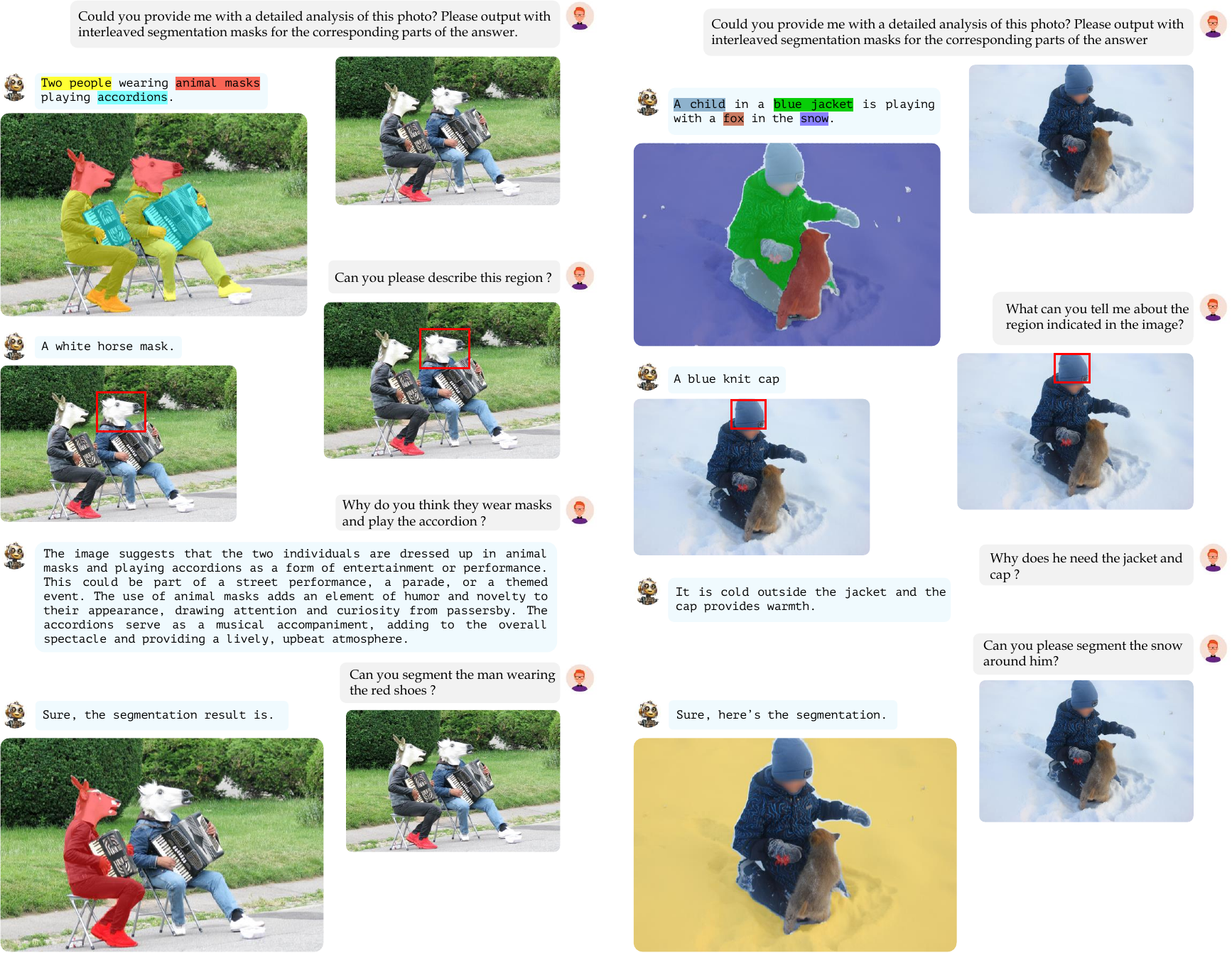}
    \caption{\textbf{Multimodal conversational with GLaMM.} The figure shows multimodal conversations generated through GLaMM. The model is flexible enough to process multimodal inputs and respond with multimodal outputs in a single conversation.}
    \label{fig:app_conv_supp}
\end{figure*}

We evaluate our model on the LLaVA-Bench~\cite{liu2023llava, liu2023improvedllava} that uses GPT-4 for evaluation of models. This benchmark tests the model on three different types of tasks: conversation question-answering, detailed descriptions, and complex reasoning tasks. The evaluation provides insights into the model's conversational and reasoning capabilities. The results in Tab.~\ref{tab:LLaVA_results} present a comparison of GLaMM with previous open-source models. 
We note that GLaMM performance is on par with the recently released LLaVA-1.5 which leverages additional data for vision-to-language alignment. 
Qualitative results are shown in Fig.~\ref{fig:conv_main} and Fig.~\ref{fig:app_conv_supp}.

\begin{table}[ht]
    \centering
    \setlength{\tabcolsep}{12pt}
    \resizebox{0.98\columnwidth}{!}{
    \begin{tabular}{llc}
        \toprule
        \textbf{Method} & \textbf{LLM} & \textbf{LLaVA\textsuperscript{W}} \\
        \midrule
        BLIP-2~\cite{li2023blip}& Vicuna-13B & 38.1 \\
        InstructBLIP~\cite{dai2023instructblip} & Vicuna-7B & 60.9 \\
        Qwen-VL~\cite{bai2023qwen}& Qwen-7B & 63.4 \\
        Qwen-VL-Chat~\cite{bai2023qwen}& Qwen-7B & 58.6 \\
        LLaVA-1.5~\cite{liu2023improved}& Vicuna-7B & 63.4 \\
\rowcolor{violet!10}        GLaMM & Vicuna-7B & \underline{63.3} \\
        \bottomrule
    \end{tabular}
    }
    \caption{\textbf{Evaluation of GLaMM on conversational style QA using LLaVA-Bench.} The table compares GLaMM's performance with previous open-source models in conversation question-answering, detailed descriptions, and complex reasoning tasks.}
    \label{tab:LLaVA_results}
\end{table}


\begin{figure*}[!th]
  \centering
    \includegraphics[width=0.99\linewidth]{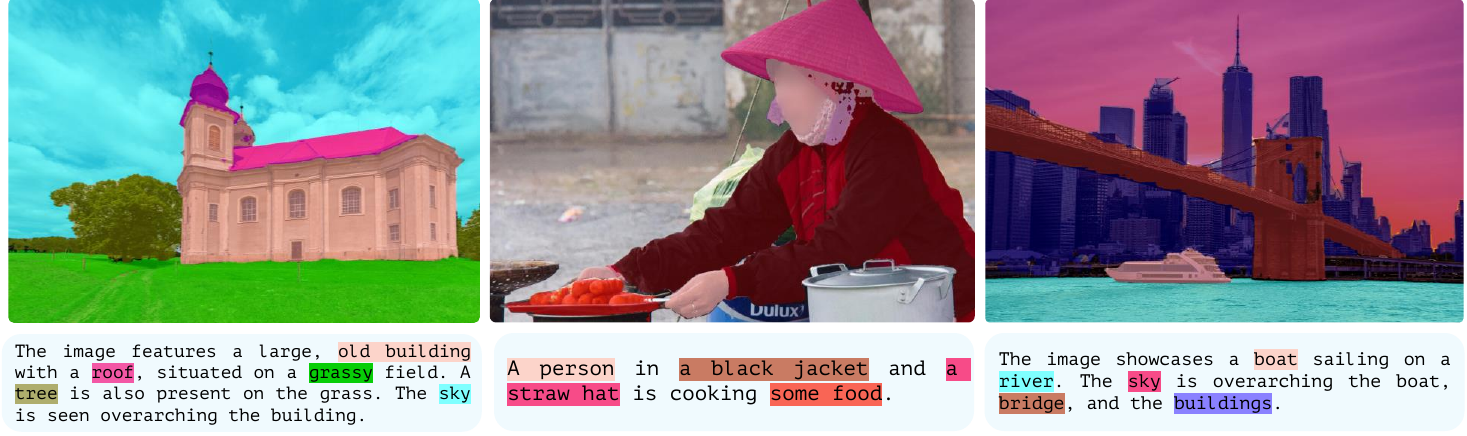}
    \caption{\textbf{Qualitative results of GLaMM's performance in grounded conversation generation.} The figure shows how GLaMM seamlessly generates detailed responses, grounding phrases using pixel-level masks showing its detailed understanding.}
    \label{fig:app_gcg2}
\end{figure*}

\begin{figure*}[!th]
  \centering
    \includegraphics[width=0.99\linewidth]{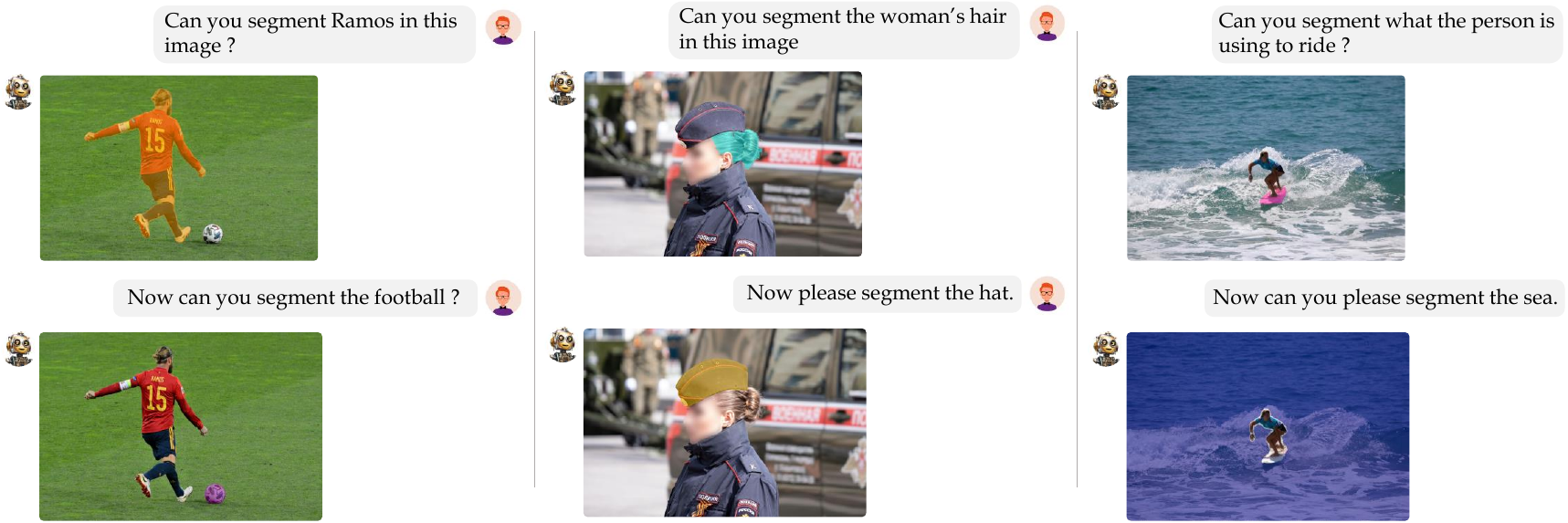}
    \caption{\textbf{Qualitative results of GLaMM's capability in referring expression segmentation.} The figure illustrates how GLaMM effectively translates text-based referring expressions into corresponding segmentation masks. Leveraging its training on the GranD dataset, the model can provide pixel-grounded reasoning and operate across various levels of granularity.}
    \label{fig:app_referseg}
\end{figure*}

\begin{figure*}[!tp]
  \centering
    \includegraphics[width=0.99\linewidth]{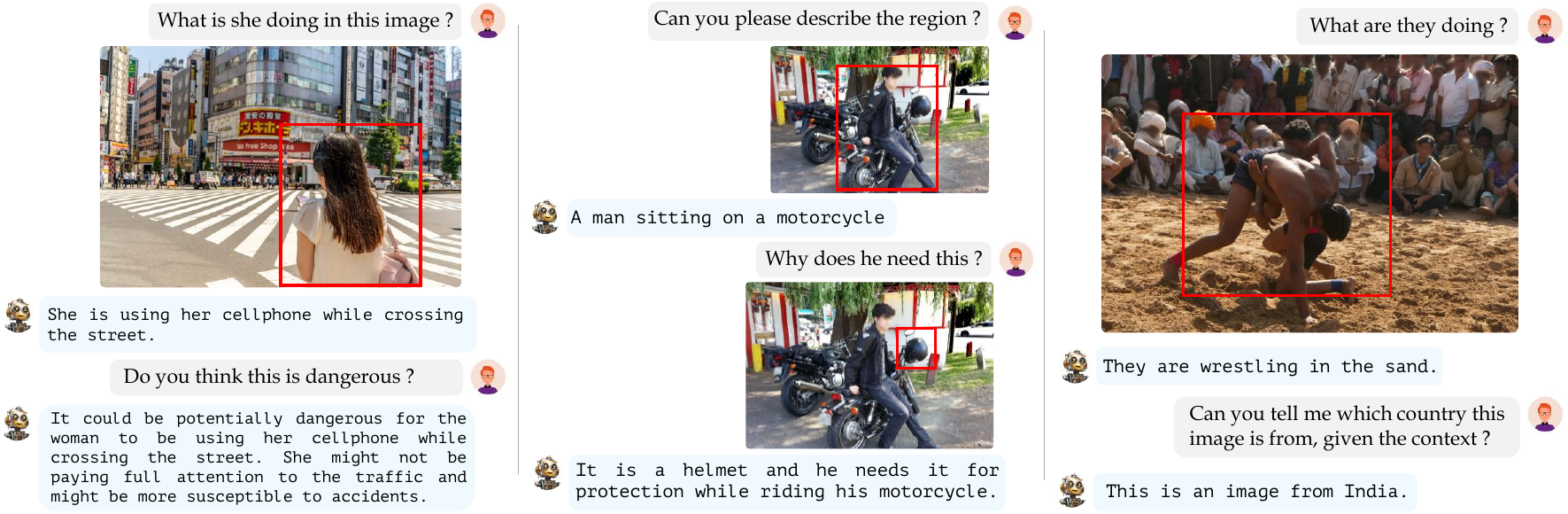}
    \caption{\textbf{Qualitative illustration of GLaMM's performance in region-level captioning.} The figure demonstrates GLaMM's ability to generate region-specific captions adeptly, translating the intricate details from designated regions into coherent textual descriptions, enriched by its training on the comprehensive GranD dataset. This capability, combined with the inherent reasoning abilities of LLMs, enables it to tackle reasoning-based visual questions about these regions.}
    \label{fig:app_regcap}
\end{figure*}

\begin{figure*}[!t]
  \centering
    \includegraphics[width=\linewidth]{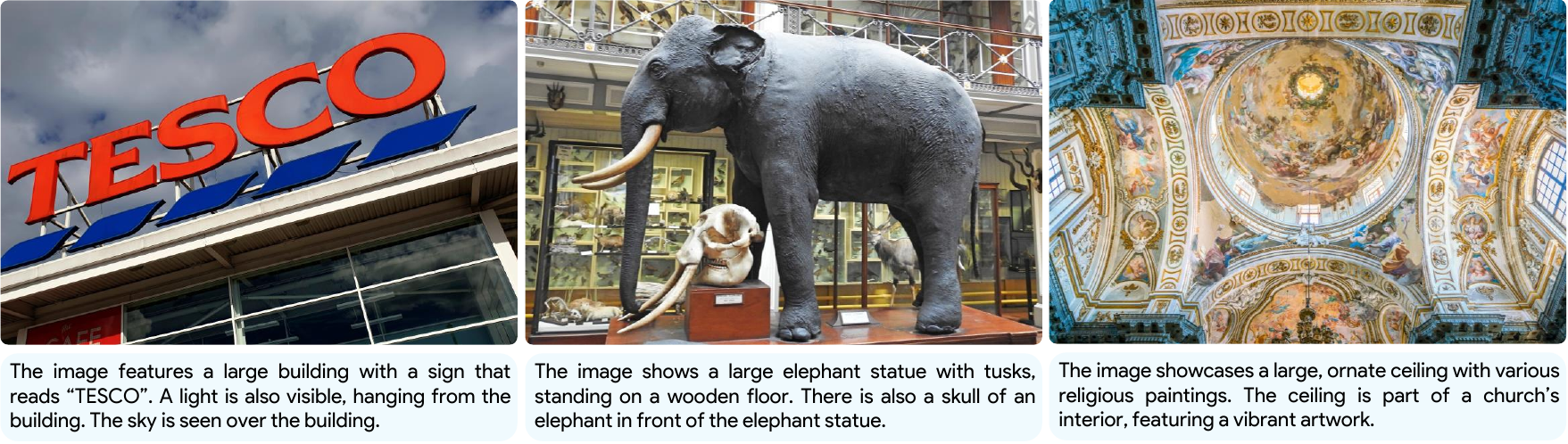}
    \caption{\textbf{Qualitative results of GLaMM on image-level captioning tasks.} The figure shows the capabilities of GLaMM in generating detailed and context-aware captions for a diverse range of images. On the left, GLaMM demonstrates its proficiency in text recognition within images; it accurately identifies and incorporates specific textual information, such as the brand name "TESCO," into its caption. In the middle image, GLaMM's capability to discern subtleties in visual content is showcased. It can effectively distinguish between live entities and inanimate objects, such as differentiating a living creature from a statue. On the right, the figure demonstrates GLaMM's competence in reasoning about complex visual scenes. It can analyze and describe intricate details and interactions within an image, reflecting a deep understanding of both the individual elements and the overall context of the scene.}
    \label{fig:app_captioning}
\end{figure*}

\begin{figure*}[!t]
  \centering
    \includegraphics[width=0.85\linewidth]{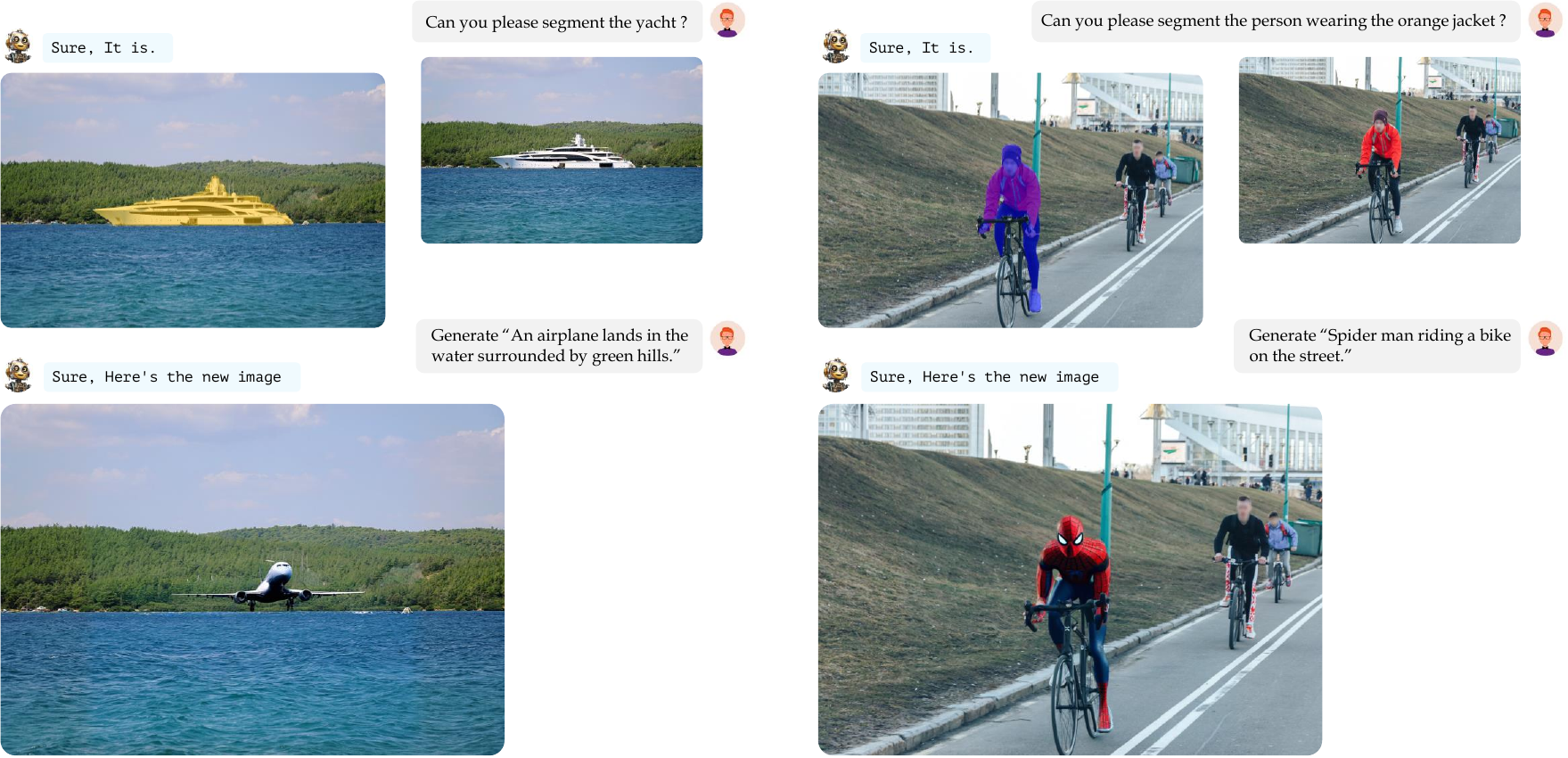}
    \caption{\textbf{Qualitative results of GLaMM on conditional image generation.} The figure shows the integration of GLaMM with an image generation model (stable diffusion). GlaMM first generates the segmentation mask (e.g. "yacht" in the left image and "person wearing orange jacket" in the right image) which is used along with a text prompt as input to the diffusion model to generate the desired images.}
    \label{fig:app_generative}
\end{figure*}

\begin{figure*}[t]
  \centering
    \includegraphics[width=0.92\linewidth]{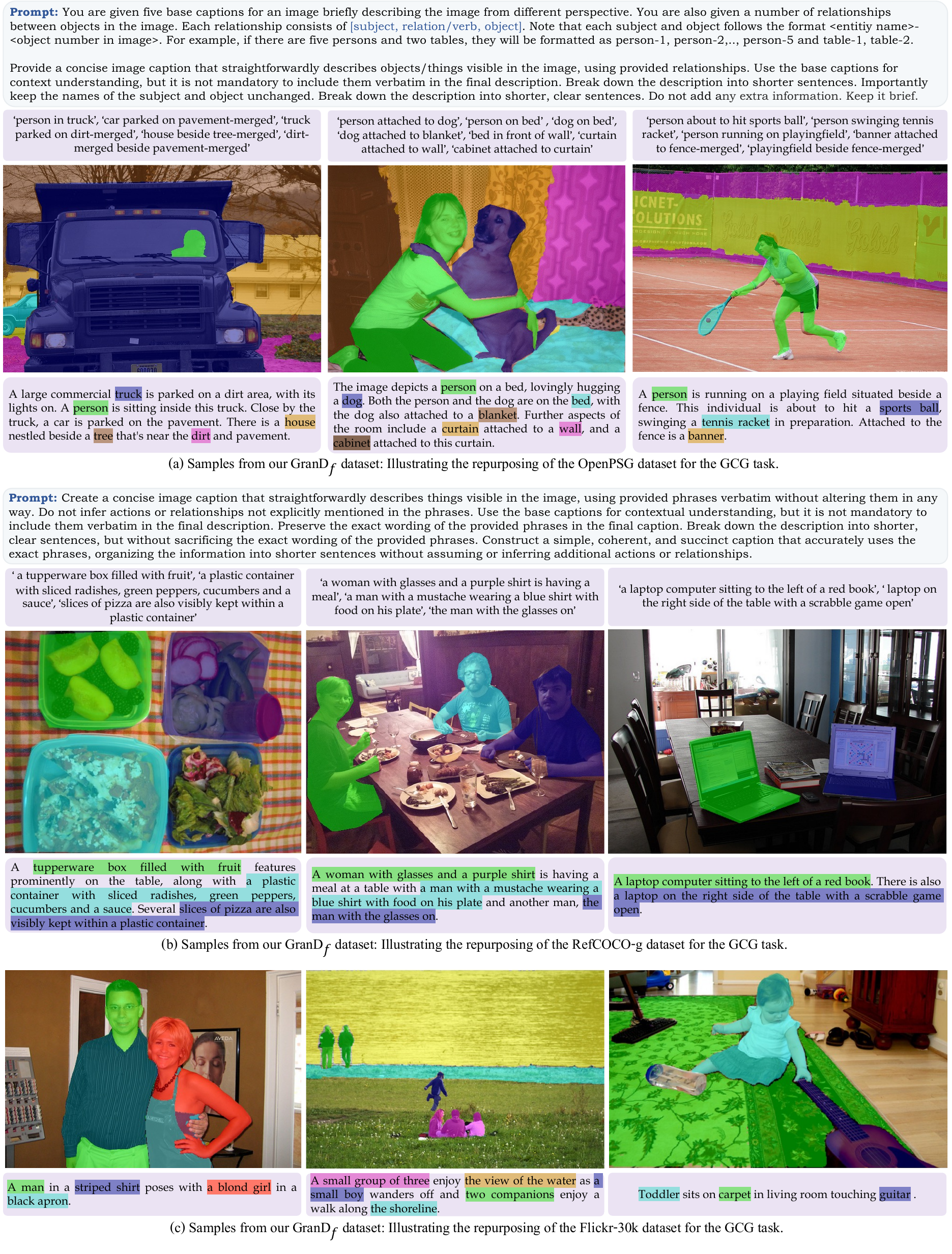}
    \caption{\textbf{Dataset samples from GranD$_f$.} The figure shows the GPT4~\cite{openai2023gpt4} prompts used and the created dataset samples from Grand$_f$ dataset. This repurposed human-annotated dataset provides rich semantics to GLaMM for GCG task.}
    \label{fig:app_dataset_gcg}
\end{figure*}

\begin{figure*}[!th]
  \centering
    \includegraphics[width=0.96\linewidth]{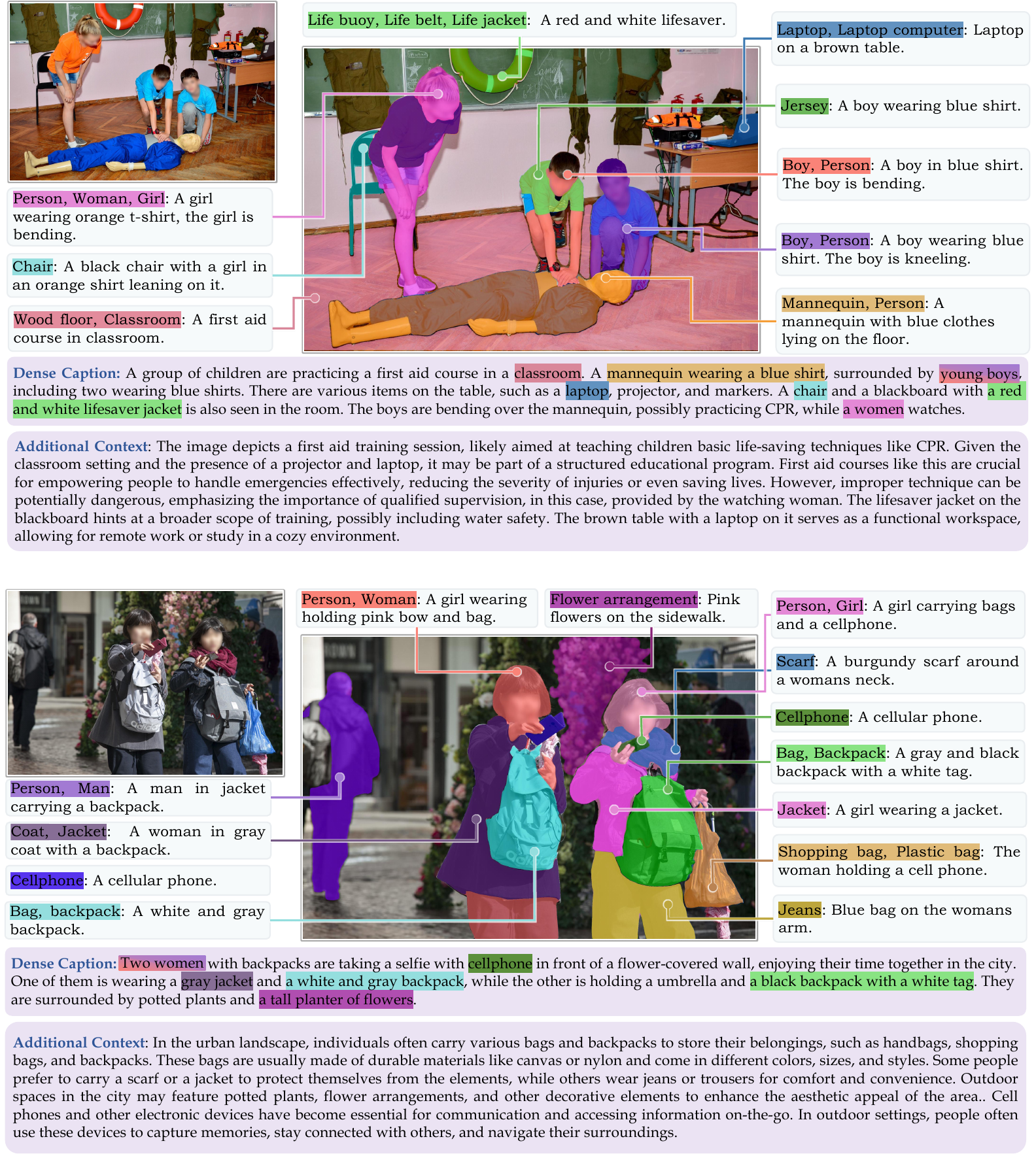}
    \caption{\textbf{Dataset samples from GranD.} The figure shows a few samples from the GranD dataset, generated using the automated annotation pipeline. It provides multiple semantic labels and attributes for detected objects, along with the grounded dense caption and additional context.}
    \label{fig:app_dataset_samples}
\end{figure*}

\section{Additional Qualitative Results}
\label{sup:additional_qualitative_results}
In this section, we provide more qualitative examples to better understand the capacity of GLaMM.

\subsection{Grounded Conversation Generation (GCG)}
Fig.~\ref{fig:app_gcg2} shows qualitative results of GLaMM finetuned on GranD$_f$ dataset. The model could produce dense captions and provide dense pixel-level groundings of the caption.

\subsection{Referring Segmentation}
Fig.~\ref{fig:app_referseg} shows the effectiveness of GLaMM in understanding the natural language query and segmenting the corresponding objects. Note that GLaMM can also segment multiple objects via multi-round conversations.

\subsection{Region-level Captioning}
Fig.~\ref{fig:app_regcap} shows the qualitative results of GLaMM for region-level understanding. Our model can generate detailed descriptions about the user-specified regions in an image.

\subsection{Image-level Captioning}
Fig.~\ref{fig:app_captioning} shows GLaMM's qualitative results on captioning tasks. Our model can generate dense captions for images.

\subsection{Conditional Image Generation}
Fig.~\ref{fig:app_generative} shows GLaMM's seamless integration for generative tasks. We use the Stable Diffusion inpainting model \textit{stable-diffusion-xl-1.0-inpainting} \cite{rombach2021highresolution} for this task. We first generate a segmentation mask using our GlaMM model based on the user query. This segmentation mask along with the user prompt is given as the input to the Stable Diffusion inpainting model, which generates the final output.

\subsection{Conversations}
Fig.~\ref{fig:app_conv_supp} illustrates the unique functionality of GLaMM to engage in multi-purpose task conversations. GLaMM is a generic conversational model that can accept prompts in the form of text and/or region and can answer in the form of text and/or segmentation masks. Note that our model is not explicitly trained to handle such scenarios, and this behavior emerges mainly due to our pretraining on GranD dataset, where an image is presented to LMM in different contexts.



\section{Dataset Visualization}
\label{sup:dataset_viz}
In this section, we provide additional dataset samples of our GranD and GranD$_f$ datasets to better understand the functionalities they offer. Please see Fig.~\ref{fig:app_dataset_samples} and Fig.~\ref{fig:app_dataset_gcg}.

\section{Limitations and Future Work}
\label{sup:limitations}
The large-scale automated pipeline provides dense labelings that are important for our pretraining but still contains some noise. A high-quality, clean dataset could help further improve the pretrained representations, although this comes at a significantly higher annotation cost. A potential research direction is to develop a cost-effective annotation pipeline aimed at reducing noise in dense labeling. Additionally, expanding the GLaMM framework to include modalities such as video and 3D is also a future research direction.

\section{Ethics and Societal Impact}
\label{sup:ethics}
Our Grounding-anything Dataset (GranD) utilizes SAM images that have de-identified personal information, with all faces and license plates obscured. To the best of our knowledge, the dataset does not portray any strong biases or discrimination. We urge for the responsible use of GranD and GLaMM, promoting research progress while safeguarding privacy.

\section{Acknowledgement}
The computations were enabled by resources provided by the National Academic Infrastructure for Supercomputing in Sweden (NAISS) at Alvis partially funded by the Swedish Research Council through grant agreement no. 2022-06725, the LUMI supercomputer hosted by CSC (Finland) and the LUMI consortium, and by the Berzelius resource provided by the Knut and Alice Wallenberg Foundation at the National Supercomputer Centre.
